\pdfoutput=1

\documentclass[11pt]{article}

\usepackage{EMNLP2023}

\usepackage{times}
\usepackage{latexsym}

\usepackage[T1]{fontenc}

\usepackage[utf8]{inputenc}

\usepackage{microtype}

\usepackage{inconsolata}

\usepackage{comment}

\usepackage{enumitem}

\usepackage{tikz}

\usepackage{ amssymb }
\usepackage{multirow}
\usepackage{array,booktabs}

\usepackage[normalem]{ulem}
\newcommand*\cloze{\tikz[baseline=(char.base)]{
            \node[shape=rectangle,minimum size=10pt,draw=lightgray, fill=black, text=white, inner sep=1pt] (char) {\tiny{C}\strut};}}
\newcommand*\questions{\tikz[baseline=(char.base)]{
            \node[shape=rectangle,minimum size=10pt,draw=lightgray, fill=black, text=white, inner sep=1pt] (char) {\tiny{Q}\strut};}}

\definecolor{aliceblue}{RGB}{102, 153, 204}
\definecolor{amber}{RGB}{238, 153, 170}
\definecolor{dkgrey}{RGB}{187, 187, 187}

\newcommand*{\cmss}{\fontfamily{lmss}\selectfont}

\newcommand {\otoprule}{\midrule [\heavyrulewidth]}
\newcolumntype {+}{ >{\global\let\currentrowstyle\relax}}
\newcolumntype {^}{ >{\currentrowstyle }}
 \newcommand {\rowstyle}[1]{\gdef\currentrowstyle{#1} %
 #1\ignorespaces
 }
\newcommand{\tabhead}{\rowstyle{\bfseries}}

\newcommand{\prex}[1]{{{\tt{#1}}}}
\newcommand*\entity{\tikz[baseline=(char.base)]{
            \node[shape=rectangle,minimum size=10pt,draw=lightgray, fill=black, text=white, inner sep=1pt] (char) {\tiny{E}\strut};}}

\newcommand*\ponone{}
\newcommand*\podm{\tikz[baseline=(char.base)]{
            \node[shape=rectangle,minimum size=10pt,draw=lightgray, fill=aliceblue, inner sep=1pt] (char) {\tiny{DM}\strut};}}
\newcommand*\podpo{\tikz[baseline=(char.base)]{
            \node[shape=rectangle,minimum size=10pt,draw=lightgray, fill=aliceblue, inner sep=1pt] (char) {\tiny{DO}\strut};}}
\newcommand*\polm{\tikz[baseline=(char.base)]{
            \node[shape=rectangle,minimum size=10pt,draw=lightgray, fill=aliceblue,inner sep=1pt] (char) {\tiny{LM}\strut};}}

\newcommand*\oonone{}
\newcommand*\oomultitoken{\tikz[baseline=(char.base)]{
            \node[shape=rectangle,minimum size=10pt,draw=lightgray, fill=amber, inner sep=1pt] (char) {\tiny{MT}\strut};}}
\newcommand*\ootyping{\tikz[baseline=(char.base)]{
            \node[shape=rectangle,minimum size=10pt,draw=lightgray,fill=amber,inner sep=1pt] (char) {\tiny{TY}\strut};}}

\newcommand*\oodebiasing{\tikz[baseline=(char.base)]{
            \node[shape=rectangle,minimum size=10pt,draw=lightgray,fill=amber,inner sep=1pt] (char) {\tiny{DEB}\strut};}}
\newcommand*\lmnoadapt{}
\newcommand*\lmfinetune{{\tiny{\checkmark}}}
\newcommand*\cls{\tikz[baseline=(char.base)]{
            \node[shape=rectangle,minimum size=10pt,draw=lightgray, fill=black, text=white, inner sep=1pt] (char) {\tiny{CLS}\strut};}}
\newcommand*\nlg{\tikz[baseline=(char.base)]{
            \node[shape=rectangle,minimum size=10pt,draw=lightgray, fill=black, text=white, inner sep=1pt] (char) {\tiny{NLG}\strut};}}

\newcommand*\ppl{\tikz[baseline=(char.base)]{
            \node[shape=rectangle,minimum size=10pt,draw=lightgray, fill=black, text=white, inner sep=1pt] (char) {\tiny{PPL}\strut};}}

\definecolor{dsgkcolor}{RGB}{64, 176, 166} 
\definecolor{dsocolor}{RGB}{225, 190, 106} 

\newcommand*\gk{\tikz[baseline=(char.base)]{
            \node[shape=rectangle,minimum size=10pt,draw=lightgray, fill=dsgkcolor, inner sep=1pt] (char) {\tiny{GK}\strut};}}
\newcommand*\dk{\tikz[baseline=(char.base)]{
            \node[shape=rectangle,minimum size=10pt,draw=lightgray, fill=dkgrey, inner sep=1pt] (char) {\tiny{DK}\strut};}}
\newcommand*\ku{\tikz[baseline=(char.base)]{
            \node[shape=rectangle,minimum size=10pt,draw=lightgray, fill=dsocolor, inner sep=1pt] (char) {\tiny{KU}\strut};}}
\newcommand*\consistency{\tikz[baseline=(char.base)]{
            \node[shape=rectangle,minimum size=10pt,draw=lightgray, fill=dsocolor, inner sep=1pt] (char) {\tiny{CO}\strut};}}
\newcommand*\kb{\tikz[baseline=(char.base)]{
            \node[shape=rectangle,minimum size=10pt,draw=lightgray, fill=dsocolor, inner sep=1pt] (char) {\tiny{KB}\strut};}}
\newcommand*\ue{\tikz[baseline=(char.base)]{
            \node[shape=rectangle,minimum size=10pt,draw=lightgray, fill=dsocolor, inner sep=1pt] (char) {\tiny{UE}\strut};}}
\newcommand*\contextk{\tikz[baseline=(char.base)]{
            \node[shape=rectangle,minimum size=10pt,draw=lightgray, fill=dsocolor, inner sep=1pt] (char) {\tiny{CK}\strut};}}  
\newcommand*\mc{\tikz[baseline=(char.base)]{
            \node[shape=rectangle,minimum size=10pt,draw=lightgray, fill=dsocolor, inner sep=1pt] (char) {\tiny{MC}\strut};}}

\newcommand{\PY}[1]{{\color{blue}#1}}
\newcommand{\CS}[1]{{\color{purple}CS: #1}}

\newif\ifshowChanges
\showChangestrue

\showChangesfalse

\ifshowChanges
    \newcommand{\CR}[1]{{\color{blue}#1}}
\else
    \newcommand{\CR}[1]{{#1}}
\fi

\title{Give Me the Facts! A Survey on Factual Knowledge Probing \\in Pre-trained Language Models}

\author{Paul Youssef\textsuperscript{\normalfont{1,3}} Osman Alperen Koraş\textsuperscript{\normalfont{1}} Meijie Li\textsuperscript{\normalfont{1}} Jörg Schlötterer\textsuperscript{\normalfont{1,2,3}} Christin Seifert\textsuperscript{\normalfont{1,3}}\\
  \textsuperscript{1}Institute for AI in Medicine (IKIM), University Hospital Essen, University of Duisburg-Essen\\ \textsuperscript{2}University of Mannheim \textsuperscript{3}University of Marburg\\
  \texttt{\{paul.youssef, joerg.schloetterer, christin.seifert\}@uni-marburg.de}\\
 \texttt{\{osman.koras, meijie.li\}@uni-due.de}}

\begin{document}
\maketitle
\begin{abstract}
Pre-trained Language Models (PLMs) are trained on vast unlabeled data, rich in world knowledge. This fact has sparked the interest of the community in quantifying the amount of factual knowledge present in PLMs, as this explains their performance on downstream tasks, and potentially justifies their use as knowledge bases. In this work, we survey methods and datasets that are used to probe PLMs for factual knowledge. Our contributions are: (1) We propose a categorization scheme for factual probing methods that is based on how their inputs, outputs and the probed PLMs are adapted; (2) We provide an overview of the datasets used for factual probing; (3) We synthesize insights about knowledge retention and prompt optimization in PLMs, analyze obstacles to adopting PLMs as knowledge bases and outline directions for future work.

\end{abstract}

\section{Introduction}
Pre-trained language models have been a game changer in NLP. Their reliance on large unlabeled corpora for pre-training and the availability of computational resources have enabled a speedy scaling of these models. This scaling has been reflected on the performance of numerous downstream tasks in NLP~\cite{devlin-etal-2019-bert, chowdhery-etal-2022-palm, touvron-etal-2023-llama}, and led to the wide adaptation of the \emph{pre-train then finetune} framework. 


The success of PLMs is attributed to the rich representations and the knowledge captured from the pre-training corpora~\cite{de-cao-etal-2021-editing, han-etal-2021-pretrained, ye-etal-2022-zerogen}. There has, therefore, been a huge interest in investigating and quantifying the type and amount of knowledge present in PLMs, e.g.,~\cite{davison-etal-2019-commonsense, jawahar-etal-2019-bert, petroni-etal-2019-language, tenney-etal-2018-context, roberts-etal-2020-much}, in order to have a better understanding about which kinds of knowledge are internalized during pre-training, and to develop methods to make PLMs more knowledge-rich and obtain gains on various downstream tasks. 

\begin{figure}
\includegraphics[scale=0.9]{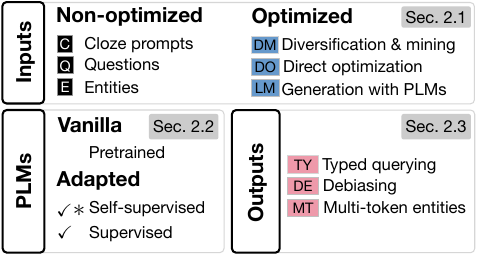}

\caption{An overview of our categorization scheme of factual knowledge probing methods.}
\label{fig:overview}
\end{figure}
Besides the interest in quantifying knowledge for better downstream tasks performance, there is a special interest in factual knowledge present in PLMs, because they are envisioned to become \textit{soft knowledge bases}, from which one can easily extract relational knowledge that had been captured during pre-training~\cite{petroni-etal-2019-language, sung-etal-2021-language}. Querying PLMs for knowledge would eliminate the complex NLP pipelines used for knowledge extraction, the need for labeled data to train models for relational knowledge extraction, and  schema designing~\cite{petroni-etal-2019-language}. Furthermore, PLMs would allow users to formulate queries to knowledge bases (KBs) in natural language, which makes them accessible to a wider user base~\cite{heinzerling-inui-2021-language}. Despite recent advances enabling smooth conversational interactions, e.g., with ChatGPT\footnote{https://openai.com/blog/chatgpt}, factuality is still an open issue~\cite{ray-2023-chatgpt}.

Many methods and datasets have been proposed to \emph{probe} PLMs for factual knowledge. Probing involves a PLM and a dataset. The dataset contains truthful facts. These facts are used to estimate the amount of knowledge in PLMs. More specifically, the dataset contains inputs that identify the fact we are looking for, in order to extract it from the PLM (e.g., ``Dante was born in \texttt{[MASK]}''), and ground truth answers that help evaluate if the retrieved answers are indeed correct (e.g., Florence). The data is often described in terms of relations (e.g., ``place-of-birth'') between subjects (e.g., ``Dante'') and objects (e.g., ``Florence''). To produce prompts, a template is created for each relation (e.g., `[X] was born in \texttt{[MASK]}''), that is then filled with subject entities. The inputs can also have other forms such as questions (e.g., ``Where was Dante born?'').  



In this work, we review recent work about factual knowledge probing. For the survey, we considered papers that cite the seminal work by~\citet{petroni-etal-2019-language} which first introduced the concept of PLMs as KBs.\footnote{For more details refer to Appendix~\ref{app:annotationsummary}} We make the following contributions: (1) We provide a categorization of factual knowledge probing methods that is based on how inputs, PLMs and their outputs are adapted (see Figure~\ref{fig:overview} and Section~\ref{sec:probing}); (2) We provide an overview of the datasets used for factual knowledge probing and categorize these under three classes based on their goal (Section~\ref{sec:datasets}); (3) We synthesize insights about knowledge retention and prompt optimization in PLMs (Section~\ref{sec:insights}), analyze obstacles to adopting PLMs as knowledge bases (Section~\ref{sec:challenges}), and outline directions for future work (Section~\ref{sec:discussion}). We make our corpus of relevant papers publicly available. 

\section{Methods for Factual Probing}

\label{sec:probing}

We categorize factual probing methods based on adaptations to i) input, ii) model, and iii) output. Categories are not mutually exclusive, i.e., one method could adapt input and model simultaneously. Figure~\ref{fig:overview} and Table~\ref{tab:probing:overview} provide an overview of the probing methods. We only consider prompting methods that have been explicitly used for factual knowledge probing. For a general review of prompting methods, we refer to~\cite{liu-etal-2023-prompt-ACM}. 




\subsection{Probing Inputs}
\label{ssec:probing-inputs}
We distinguish between non-optimized or fixed inputs, and optimized inputs that are adapted in various ways to elicit more facts from PLMs.  
\subsubsection{Non-optimized Inputs} 
Extracting factual knowledge from PLMs depends on providing them with short inputs that indirectly describe the sought-after information. These methods can take various forms (cloze prompts~\cite{taylor-1953-cloze}, questions, or entities). Non-optimized inputs represent the simplest case, where the probing inputs are not altered in any way. 


\paragraph{Cloze prompts} are widely used across several methods. \citet{petroni-etal-2019-language} probe PLMs for factual knowledge by manually constructing cloze-style templates for several  relations. \citet{onoe-etal-2022-entity} automatically construct cloze prompts from Wikipedia and Wikidata by masking out spans near entities of interest, in order to evaluate PLMs' knowledge about (unseen) entities. \citet{abaho-etal-2022-position} construct cloze prompts from annotated PubMed abstracts to use PLMs as health outcome predictors. \citet{chen-etal-2022-meta} finetune PLMs using cloze prompts that consist of task descriptions alongside a few examples to elicit more facts. 

\paragraph{Questions} are the second input category. Several Question Answering datasets are used to finetune T5 models~\cite{raffel-etal-2020-exploring}, and evaluate the amount of knowledge implicitly present in their parameters  in \cite{roberts-etal-2020-much}. Multiple choice questions are used in~\cite{hardalov-etal-2020-exams} by providing PLMs with the questions followed by each option individually. The options are masked, and the final answer is selected based on the normalized log probabilities of the predicted tokens for each option. \citet{kalo-fichtel-2022-kamel} present a dataset based on Wikipedia, where inputs consist of several questions and answers, i.e., a few examples to implicitly indicate the task, and a similar question without an answer for evaluation. 

\paragraph{Entities} are used in methods that infer relational information or generate descriptions based on these entities. Some methods depend on a simple classifier or cosine similarity between the subject and object representations to determine the presence or absence of a relation. For example, to probe for geographical knowledge,~\citet{lietard-etal-2021-language} use fixed inputs that contain locations (e.g., countries or cities). These inputs are then used to extract representations for the respective locations from PLMs. Using these representations, the authors evaluate based on the ability of a simple classifier to solve certain tasks (e.g., predicting if two countries share border). \citet{dufter-etal-2021-static} evaluate the amount of knowledge present in static word embeddings by matching a subject entity (the query) to an object entity from a pre-defined set of possible objects based on the cosine similarity between the representations of the subject and object entities. \citet{shi-etal-2021-descgen} train generative PLMs to generate entities' descriptions while providing only the entities as inputs, and compare them to ground truth descriptions. 

\subsubsection{Optimized Inputs} 
Probing inputs contribute substantially to the probing procedure. PLMs are sensitive to the inputs~\cite{petroni-etal-2019-language,jiang-etal-2020-know, elazar-etal-2021-measuring}, and even syntactical variations or distractors, that do not alter the meaning, cause the PLM's predictions to change~\cite{heinzerling-inui-2021-language, longpre-etal-2021-entity,  pandia-ettinger-2021-sorting, podkorytov-etal-2021-limitations, li-etal-2022-eliteplm}. Therefore, depending on the probing inputs, the estimate on factual knowledge we obtain may vary significantly. Optimized inputs represent variations of the inputs, where the inputs are changed to account for the sensitivity of the probed PLMs. 

\paragraph{Diversification and mining} methods aim to diversify and optimize prompts by mining Wikipedia or other resources, and selecting the best performing prompts or a combination of them. For example, \citet{jiang-etal-2020-know} propose a mining-based and a paraphrasing-based approach to create alternative prompts that outperform manual ones. The final prompts are selected based on their performance on a training set, and can also be combined in an ensemble. \citet{bouraoui-etal-2020-inducing} mine for prompts that contain the entities of interest, and filter these based on the ability of the probed PLMs to predict the masked objects. After the filtering step, the remaining prompts are utilized to create a dataset that consists of positive inputs, i.e., containing true subject-object pairs, and negative inputs, which contain false pairs. This dataset is then used for the final evaluation. 

\paragraph{Direct optimization} methods aim to directly optimize existing prompts. This optimization happens either in a discrete space, to keep the prompts in natural language, or in a continuous space where the prompts do not have to correspond to specific tokens from the vocabulary. Optimization could also target only the masked token or the order of the examples in the prompt, in case a few examples are provided in the prompt to better indicate the task. \citet{shin-etal-2020-autoprompt}'s AUTOPROMPT extends manually created prompts by prompts with a pre-defined number of trigger tokens, and employs gradient-based search to sequentially replace the trigger tokens with concrete tokens. These tokens are chosen to increase the probability of predicting the correct object. OPTIPROMPT~\cite{zhong-etal-2021-factual} is similar to AUTOPROMPT, but allows for the trigger tokens to be replaced with vectors from a continuous embedding space. In a similar fashion,~\citet{qin-eisner-2021-learning} propose learning an ensemble of continuous prompts per relation. Additionally, they perturb the representations of the prompts in each layer in the probed PLMs using small learnable vectors. The intuition is to have activation patterns that are similar to the ones encountered during pre-training, which would make it easier to elicit knowledge from PLMs. \citet{newman-etal-2022-padapters} utilize adapters~\cite{houlsby-etal-2019-adapters} to map the embedding vectors to continuous prompts in order to make the probed PLMs less sensitive to different phrasings of the same prompts. \citet{saeed-papotti-2022-type} augment the masked tokens with a special type of embeddings, called Type Embeddings. These embeddings are derived from several entities that share the same type, and are shown to help tie the probed PLM's predictions to the expected type of the masked entity. PERO~\cite{kumar-talukdar-2021-reordering} depends on querying PLMs with prompts containing few training examples (or shots), which demonstrate the task to the queried PLMs. Since PLMs are quite sensitive to the order and the quality of the provided training examples in the prompt, PERO leverages a genetic algorithm to find an optimized prompt and a separator token to concatenate the examples in the prompts. \CR{ \cite{li-etal-2022-spe} exploit the symmetry of the task, and optimize prompts in a continuous space so that the probability of predicting both the subject and the object is maximized using the resulting prompts.}
\paragraph{Generation with PLM} methods re-write prompts with the help of a secondary PLM. \citet{haviv-etal-2021-bertese} re-write manual prompts using another version of the probed model. The re-writing model is trained to produce prompts that help extract more knowledge from the probed one, which is kept unchanged. \citet{zhang-etal-2022-promptgen} leverage a generative PLM to produce optimized prompts. 
\subsection{Probed PLMs}
\label{ssec:probed-plms}
PLMs are probed for knowledge using either their original pre-trained parameters~\cite{petroni-etal-2019-language, jiang-etal-2020-know}, or after adapting these parameters~\cite{roberts-etal-2020-much, meng-etal-2022-rewire}.

\subsubsection{Vanilla PLMs} 
Methods in this category do not induce any changes to the probed PLMs, and depend on pre-training objectives to probe PLMs for factual knowledge. Using the pre-trained parameters is the most straightforward approach and is claimed to preserve the facts learned during pre-training~\cite{elazar-etal-2021-measuring, newman-etal-2022-padapters}.

Most methods leverage the language modeling objectives from pre-training to probe for factual knowledge~\cite{petroni-etal-2019-language, jiang-etal-2020-know, shin-etal-2020-autoprompt, haviv-etal-2021-bertese, kumar-talukdar-2021-reordering, zhong-etal-2021-factual, kalo-fichtel-2022-kamel, newman-etal-2022-padapters, onoe-etal-2022-entity,saeed-papotti-2022-type}. Other methods rely on representations that come from the model's body, discarding task-specific parameters altogether (e.g., the Masked Language Modeling head in BERT-like models)~\cite{lietard-etal-2021-language} or use representations of the subject and object entities in the case of static word embeddings~\cite{dufter-etal-2021-static}.



\subsubsection{Adapted PLMs} 
Some works adapt the PLMs under evaluation to enable evaluation tasks, that do not correspond to any pre-training objective. The adaptation, however, is also coupled with risks such as train-test overlap~\cite{lewis-etal-2021-question, wang-etal-2021-generative}.
\paragraph{Supervised adaptation.} Most methods finetune the probed PLMs in a supervised manner to adapt them to the probing task. \citet{roberts-etal-2020-much} finetune T5 models for closed-book question answering, where models have only questions as inputs, while leaving out any context or external knowledge sources that might contain the answer. Similarly,~\citet{wang-etal-2021-generative} finetune BART to output a related passage, and then the answer. \citet{bouraoui-etal-2020-inducing} finetune BERT to classify prompts based on whether the relation between the subject and object entities truly holds or not. \citet{fichtel-etal-2021-prompt} finetune a BERT model with its masked language modeling head to predict the masked tokens in the provided prompts. \citet{abaho-etal-2022-position} propose an additional position-attention layer on top of transformer models, where the position of the masked token is kept constant, and the remaining tokens are given positions relative to the masked token. This approach is considered to put more focus on the masked tokens and its interaction with the remaining tokens in the prompt. 
\citet{chen-etal-2022-meta} leverage a task description that depends on the relation between the subject and object entity, alongside a few labeled examples to train the probed PLMs. At inference time, the PLMs are kept frozen and are provided with unseen task descriptions and labeled examples to adapt to the task. \citet{elazar-etal-2021-measuring} further train BERT with a consistency loss to increase its robustness to paraphrases that describe the same relation. \citet{shi-etal-2021-descgen} finetune generative PLMs to generate entity descriptions depending only on their knoweldge from pre-training. \citet{qin-eisner-2021-learning} do not directly change any parameters in PLMs, but rather introduce additional trainable parameters in each layer that change the hidden representations of the prompts to help make them more suitable for knowledge extraction. 

\paragraph{Self-supervised adaptation.} Adaptations in a self-supervised manner can introduce changes to the model without explicitly finetuning the model to the probing task. For example, \citet{meng-etal-2022-rewire} propose to \emph{re-wire} the probed PLM in a self-supervised manner. Their method depends on using data from the pre-training phase, splitting each sentence into a head part and a tail part, and using a contrastive learning objective to push the representations of the matching head and tail pairs (positives) closer to one another, and that of the non-matching pairs (negatives) to be further apart. The evaluation is based on the similarity between the representations of the prompt and a predefined set of entities that represent potential answers.

\subsection{Outputs} 
\label{ssec:outputs}

Methods focusing on the outputs of PLMs address restricting the output space of PLMs, debiasing their outputs, and handling multi-token entities.

\paragraph{Typed querying.}~\citet{kassner-etal-2021-multilingual} propose to restrict the space of possible values for replacing the masked token (object) from the whole vocabulary to a specific set of tokens whose type matches the type of the ground truth object. For example, if the PLM is queried with the prompt: ``The smallest country in the world is \texttt{[MASK]}'', only entities of type country are considered to replace the \texttt{[MASK]} token. This method has two advantages: it reduces the number of objects under consideration and allows for a better comparison across PLMs with different vocabularies~\cite{kassner-etal-2021-multilingual}. 

\paragraph{Debiasing.} \citet{zhao-etal-2021-calibrate} identify biases in the predictions of PLMs towards common and recent tokens, and propose a method that adapts the output probabilities by first estimating these biases using neutral examples and then correcting them. This debiasing method is shown to reduce the variance across prompts and has a positive effect on fact retrieval. \citet{malkin-etal-2022-coherence} propose a method to increase the effect of distant tokens on the predictions of PLMs. The method depends on combining two output distributions over the vocabulary. One distribution is based on the full-length input, whereas the other is based on a shortened version of the same input. \CR{\citet{wang-etal-2023-towards-alleviating} identify the problem of object bias in optimized prompts and propose to make all potential objects equally probable when no subject is provided, and increasing the probability of the correct object, when the subject is available. \citet{yoshikawa-okazaki-2023-selective} output predictions only above a sufficient confidence threshold. This results in a less biased evaluation, and reflects the ability of PLMs in excluding uncertain predictions. To address the problems of multiple valid answers and frequency bias, i.e., the co-occurence of some subject and object entities despite not being in a factual relation to one another, \citet{dong-etal-2022-calibrating} use two templates, one contains the correct relation while the other contains an erroneous relation between the two entities, and compare the probability for the correct object under both relations. }

\paragraph{Multi-token entities.} To handle multi-token entities,~\citet{jiang-etal-2020-x} propose using a pre-defined number of masked tokens and filling these using different strategies: 1) independent from each other, 2) sequentially (left-to-right for English), 3) starting with the most confident predictions. \CR{\cite{kalinsky-etal-2023-simple} leverage the masked token representation to generate multiple tokens using a small generative model. }



\begin{table*}
\begin{minipage}{\linewidth}
\scriptsize
\centering

\begin{tabular}{+>{\raggedright}p{2.8cm}^c^c^c^p{5cm}^>{\raggedright}p{2.6cm}^p{1.2cm}} 

\toprule\tabhead
Paper & \rotatebox[origin=c]{90}{Input} & \rotatebox[origin=c]{90}{Opt.} & \rotatebox[origin=c]{90}{Adapt.} & Example & Tested PLMs & Eval. \\\otoprule
\citet{petroni-etal-2019-language} 
    & \cloze
    & \ponone\oonone
    & \lmnoadapt
    & \prex{Dante was born in [MASK].} 
    & fairseq-fconv,  ELMo, Transformer-XL,  BERT
    & p@k\\\midrule
\citet{bouraoui-etal-2020-inducing} 
    & \cloze
    & \podm\oonone
    & \lmfinetune
    & mining trigger prompts \newline \prex{[X] is the capital of [Y].}
    & BERT
    & F1 \\
\citet{hardalov-etal-2020-exams}
    & \questions	
    & \ponone\oonone
    & \lmnoadapt+\lmfinetune\footnote{adapted and non-adapted PLM}
    & \prex{<Q> $\rightarrow$ <A1,A2,A3,A4>}	
    & XLM-R	
    & p@1	\\
\citet{jiang-etal-2020-x}
    & \cloze
    & \ponone\oomultitoken
    & \lmnoadapt
    & \prex{Barack Obama is a [MASK] [MASK] [MASK] by profession.}
    & mBERT, XLM, XLM-R	
    & p@1 \\														
\citet{jiang-etal-2020-know}
    & \cloze	
    & \podm\oonone	
    & \lmnoadapt
    & prompt mining and paraphrasing \newline \prex{DirectX is developed by [MASK]. [MASK] released DirectX. DirectX is created by [MASK].}
    & BERT, ERNIE, KnowBert	
    & p@1 \\
\citet{roberts-etal-2020-much}
    & \questions
    & \ponone\oonone
    & \lmfinetune
    & \prex{Who lives in the imperial palace in Tokyo?}	
    & T5
    & EM \\
\citet{shin-etal-2020-autoprompt}
    & \cloze	
    & \podpo\oonone	 
    & \lmnoadapt
    & prompts optimization in discrete space \newline \prex{[X] is memory arcade branding by [MASK]}
    & BERT, RoBERTa
    & p@1, p@10, MRR \\
\midrule				
\citet{dufter-etal-2021-static} 
    & \entity
    & \ponone\oonone
    & \lmnoadapt
    & \prex{sim(<capital entity>, <country entity>)}	
    & BERT, mBERT, fastText
    & p@1\\

\citet{elazar-etal-2021-measuring} 
    & \cloze
    & \ponone
    & \lmfinetune
    & trains PLM with consistency loss \newline \prex{The capital of Italy is [MASK], Italy’s capital, [MASK].}
    & BERT
    & p@1, cons, cacc\\
    
\citet{fichtel-etal-2021-prompt} 
    & \cloze	
    & \ponone\oonone
    & \lmfinetune
    & \prex{Dante was born in [MASK].} 	
    & BERT	
    & p@1 \\
\citet{haviv-etal-2021-bertese}	
    & \cloze
    & \polm\oonone
    & \lmnoadapt
    & re-writing with PLM \newline \prex{will \& grace is originally aired on [MASK].}	
    & BERT	
    & p@1\\
\citet{kassner-etal-2021-multilingual}
    & \cloze	
    & \ponone\ootyping
    & \lmnoadapt
    & \prex{Berlin is the capital of [MASK]\textsubscript{country}} 
    & BERT, mBERT
    & p@1 \\
\citet{kumar-talukdar-2021-reordering}
    & \cloze
    & \podpo\oonone	
    & \lmnoadapt
    & examples reordering \newline \prex{ex1, ex2, ex3, Rome is located in [MASK].}
    & BERT
    & p@1 \\
\citet{lietard-etal-2021-language}
    & \entity
    & \ponone\oonone
    & \lmnoadapt
    & \prex{He lives in <location entity>.}	
    & BERT, RoBERTa, GPT-2	
    & PER \\
\citet{qin-eisner-2021-learning}
    & \cloze
    & \podpo\oonone
    & \lmfinetune
    & prompts optimization in continuous space, perturbations of representations in all layers \newline \prex{[X] [V1] $\ldots$  [V5] [MASK] [V6]}	
    & BERT, RoBERTa
    & p@1, p@10, MRR \\
    
\citet{shi-etal-2021-descgen}	
    & \entity	
    & \ponone\oonone
    & \lmfinetune
    & generating descriptions for entities \newline \prex{[Carl Menger] was an Austrian...}	
    & BART, T5
    & R-L \\		

\citet{wang-etal-2021-generative}
   & \questions
    & \ponone\oonone

   & \lmfinetune
   & \prex{<Q> $\rightarrow$ <answer related passage> <A>}
   & BART
   & EM, F1, HE \\

\citet{zhao-etal-2021-calibrate}	
    & \cloze
    & \oodebiasing
    & \lmnoadapt
    & estimates and corrects biases \newline \prex{NA was born in [MASK].} 
    & GPT-3
    & p@1 \\
    
\citet{zhong-etal-2021-factual}	
    & \cloze
    & \podpo\oonone
    & \lmnoadapt
    & prompts optimization in continuous space \newline \prex{ [X] [V1] $\ldots$ [V5]  [MASK]}	
    & BERT	
    & p@1 \\
    \midrule
\citet{abaho-etal-2022-position}
    & \cloze
    & \ponone\oonone
    & \lmfinetune
    & \prex{Two CMZ patients and one morphine patient showed complete [MASK].}	
    & BERT, BioBERT, Biomed\_RoBERTa, SciBERT, UmlsBERT
    & EM, PM \\
\citet{chen-etal-2022-meta}	
    & \cloze
    & \ponone\oonone
    & \lmfinetune
    & \prex{<task description> <example>* Dante was born in [MASK].}
    & BERT, DeBERTa	
    & p@1 \\

\CR{ \citet{dong-etal-2022-calibrating} }
    & \cloze
    & \oodebiasing 
    & \lmnoadapt
    & uses probabilities for correct/incorrect relations \CR{\prex{P(Hawaii | Obama was born in) / P(Hawaii | Obama worked)} } 
    & T5
    & False rate\\
    
\citet{kalo-fichtel-2022-kamel} 
    & \questions
    & \ponone\oonone
    & \lmnoadapt
    & \prex{<Q\&A>*, What languages does Confuzius speak?} 
    & GPT-J, GPT-2, OPT	
    & F1 \\

\CR{\citet{li-etal-2022-spe} }
    & \cloze
    & \podpo\oonone	
    & \lmnoadapt
    & \CR{optimized prompts to predict subject and object \prex{([MASK]) [V1] $\ldots$ [V5]  ([MASK])} }
    & BERT, RoBERTa
    & p@k, MRR \\
    
\citet{malkin-etal-2022-coherence} 
    & \cloze
    & \oodebiasing 
    & \lmnoadapt
    & combines two output distributions \newline \prex{Dante was born in [MASK], was born in [MASK]} 
    & GPT-2, GPT-3
    & p@1	\\
    
\citet{meng-etal-2022-rewire} 
    & \cloze
    & \ponone\oonone
    & \lmfinetune $\ast$
    & \prex{sim(Elvitegravir may prevent [MASK], entity) }	
    & BERT, BlueBERT, BioBERT, T5, BART, PubMedBERT, SciFive
    & p@1	\\
\citet{newman-etal-2022-padapters}	
    & \cloze
    & \podpo\oonone
    & \lmnoadapt
    & adapter mapping prompts to continuous prompts after embedding layer \newline \prex{[V1] $\ldots$  [V5]} from \prex{[MASK] is Canada's capital}
    & BERT	
    & p@1, cons\\	
\citet{onoe-etal-2022-entity} 
    & \cloze	
    & \ponone\oonone
    & \lmnoadapt
    & \prex{[mRNA vaccines] do not affect [MASK].} 	
    & T5, BART, GPT-Neo	
    & pplx\\
\citet{saeed-papotti-2022-type}	
    & \cloze	
    & \podpo\oonone
    & \lmnoadapt
    & masked tokens with type embeddings \newline \prex{The wife of Obama is ([MASK] + [TE]).}	
    & BERT
    & p@1, p@k	\\		
\citet{zhang-etal-2022-promptgen} 
    & \cloze
    & \polm\oonone
    & \lmnoadapt
    & generating prompts by PLM (BART) \newline \prex{Marco Benevento and not violin yeah much like trafficking UNESCO partly [MASK].}	
    & BERT
    & p@1\\									
    \midrule

\CR{\citet{kalinsky-etal-2023-simple}}
    & \cloze
    & \ponone\oomultitoken
    & \lmnoadapt
    & \CR{uses the masked token repr. to generate multi-token predictions \newline\prex{I love [MASK] city.}}
    & BERT
    & p@1 \\

\CR{ \citet{wang-etal-2023-towards-alleviating} }
    & \cloze
    & \oodebiasing 
    & \lmnoadapt
    & \CR{reduces object bias \newline\prex{The native language of [X] is [MASK].} }
    & BERT, RoBERTa
    & p@1, MRR, entropy	\\

\CR{ \citet{yoshikawa-okazaki-2023-selective} }
    & \cloze
    & \oodebiasing 
    & \lmnoadapt
    & \CR{outputs prediction by sufficient confidence \newline \prex{[X] was born in [MASK].} }
    & BERT, RoBERTa
    & p@1, RC-AUC\\
    \bottomrule
\end{tabular}

\caption{Overview of probing methods. Input type: cloze prompts \cloze, questions \questions, entities \entity. Prompt optimization: diversification and mining \podm, direct  optimization \podpo, or generation with PLMs \polm. Other methods: debiasing \oodebiasing, mutli-token entities \oomultitoken, or typed querying \ootyping. PLM adaptation: supervised (\lmfinetune), or self-supervised (\lmfinetune $\ast$). Evaluation: consistency (cons), consistent-accuracy (cacc), exact match (EM), human evaluation (HE), mean reciprocal rank (MRR), partial match (PM), perplexity (pplx), probe error reduction (PER), ROUGE-L (R-L), and AUC of the risk-coverage curve (RC-AUC).}

\label{tab:probing:overview}
\end{minipage}
\end{table*}

\section{Datasets for Factual Probing}
\label{sec:datasets}


\begin{table*}
\begin{minipage}{\linewidth}
\scriptsize
\centering
\begin{tabular}{+l^>{\raggedright\hangindent=0.5em}p{3.9cm}^l^p{0.6cm}^p{6.1cm}^c^c}

\toprule \tabhead
& Dataset &  Cat. &   Lang. &  Example & \#Inst.  & Access \\
\otoprule
\multirow{25}{*}{\rotatebox[origin=c]{90}{\sc General knowledge}}
& {\cmss LAMA}~\cite{petroni-etal-2019-language} 
    & \gk \cloze
    & en  
    & \prex{Dante was born in [MASK]} 
    & 40k  
    & + \\ 
& {\cmss Google Analogy(semantic)} \newline\cite{bouraoui-etal-2020-inducing} 
    & \gk \cls 
    & en 
    &  \prex{It is located in [X], the capital of [Y]} 
    & 9k 
    & + \\                       
& {\cmss WebQuestions}~\cite{roberts-etal-2020-much} 
    & \gk \questions 
    & en  
    & \prex{What degrees did Obama get?} 
    & 6k 
    & + \\         
& {\cmss BATS (ency.)}~\cite{bouraoui-etal-2020-inducing} 
    & \gk \cls 
    & en 
    & \prex{[X] is the capital of [Y]} 
    & 0.5k 
    & + \\                            
& {\cmss TriviaQA}~\cite{roberts-etal-2020-much} 
    & \gk \questions 
    & en  
    & \prex{Who won the Nobel Peace Prize in 2009?} 
    & 96k 
    & + \\              
& {\cmss NQ}~\cite{roberts-etal-2020-much} 
    & \gk \questions
    & en  
    & \prex{Who lives in the imperial palace in Tokyo?} 
    & 322k 
    & + \\
& {\cmss IndicGLUE}~\cite{kakwani-etal-2020-indicnlpsuite} 
    & \gk \cloze 
    & indic\footnote{11 different languages}
    & \prex{Shambhupara <MASK> is an important village in Amreli Tehsil, Gujarat State.}
    & 239k 
    & + \\ 
& {\cmss X-FACTR}~\cite{jiang-etal-2020-x} 
    & \gk \cloze
    & multi 
    & \prex{The mother tongue of Obama is [MASK]} 
    & 398k 
    & + \\
& {\cmss LAMA-UHN}~\cite{poerner-etal-2020-e}    
    & \gk \cloze 
    & en  
    & \prex{USA maintains diplomatic relations with [MASK]} 
    & 32k  
    & o \\ 
& {\cmss LPAQA}~\cite{jiang-etal-2020-know} 
    & \gk \cloze
    & en 
    & \prex{DirectX is developed/created by [MASK]} 
    & 3k 
    & + \\ 
& {\cmss mLAMA}~\cite{kassner-etal-2021-multilingual} 
    & \gk \cloze 
    & multi 
    & \prex{Paris is the capital of [MASK]} 
    & 855k 
    & + \\
& {\cmss DESCGEN}~\cite{shi-etal-2021-descgen} 
    & \gk \nlg 
    & en 
    & \prex{[Carl Menger] was an Austrian economist...} 
    & 37k 
    & + \\
& {\cmss WIKI-UNI}~\cite{cao-etal-2021-knowledgeable} 
    & \gk \cloze 
    & en 
    & \prex{Turing was born in [MASK].} 
    & 70k 
    & + \\
& {\cmss SQuAD}~\cite{wang-etal-2021-generative} 
   & \gk \questions
   & en
   & \prex{<Q> $\rightarrow$ <answer related passage> <A>}
   & 92k
   & + \\
   
& {\cmss KAMEL}~\cite{kalo-fichtel-2022-kamel} 
    & \gk \questions 
    & en 
    & \prex{<Q\&A>*, What languages does Confuzius speak?} 
    & 47k 
    & + \\  

& \CR{{\cmss DLAMA}~\cite{keleg-magdy-2023-dlama}}
    & \gk \cloze
    & multi
    & \prex{Egypt is located in [MASK]} 
    & 78k
    & + \\
& \CR{{\cmss PopQA}~\cite{mallen-etal-2023-trust}}
    & \gk \questions 
    & en 
    & \prex{What is the capital of Louisiana?} 
    & 14K 
    & + \\

& \CR{{\cmss EntityQuestions}~\cite{mallen-etal-2023-trust}}
    & \gk \questions 
    & en 
    & \prex{Who is the author of The Target?} 
    & 177k
    & + \\
    \midrule
\multirow{10}{*}{\rotatebox[origin=c]{90}{\sc Domain-specific}}
& {\cmss EXAMS}~\cite{hardalov-etal-2020-exams} 
    & \dk \questions
    & multi 
    & \prex{<Q> <A1,A2,A3,A4> $\rightarrow$ <A\textsubscript{i}>}	 
    & 24k 
    & + \\
& {\cmss MedQA}~\cite{jin-etal-2021-disease} 
    & \dk \questions
    & en,zh\footnote{including zh-simplified}
    &  \prex{<Case> <Q> <A1,A2,A3,A4> $\rightarrow$ <A\textsubscript{i}>} 
    & 61k 
    & + \\             
& {\cmss DisKnE}~\cite{alghanmi-etal-2021-probing} 
    & \dk \cls
    & en 
    & \prex{The patient has high BP <SEP> Hypertension} 
    & 7k 
    & o \\  
& \cite{yuan-etal-2021-improving} 
    & \dk \cloze 
    & en 
    & \prex{apraclonidine may prevent [MASK]} 
    & 144k 
    & o \\

& {\cmss LEFT}~\cite{ciosici-etal-2021-perhaps} 
    & \dk \cls 
    & en 
    &  \prex{<statement> $\rightarrow$ <True/False>} 
    & 1k 
    & o \\

& {\cmss BioLAMA}~\cite{sung-etal-2021-language} 
    & \dk \cloze
    & en  
    & \prex{Hepatitis has symptoms such as [MASK]} 
    & 49k 
    & + \\ 
&  {\cmss EBM-NLP}~\cite{abaho-etal-2022-position}
    & \dk  \cloze
    & en  
    & \prex{...patient showed complete [MASK]} 
    & 3k 
    & - \\ 
& {\cmss MedLAMA}~\cite{meng-etal-2022-rewire} 
    & \dk \cloze  
    & en  
    & \prex{Elvitegravir may prevent [MASK]} 
    & 19k 
    & + \\  \midrule
\multirow{28}{*}{\rotatebox[origin=c]{90}{\sc Other}}
& {\cmss Negated LAMA}\newline\cite{kassner-schutze-2020-negated} 
    & \consistency \cloze
    & en 
    & \prex{The capital of Italy is not [MASK]} 
    & 10k 
    & + \\
& {\cmss Misprimed LAMA}\newline \cite{kassner-schutze-2020-negated} 
    & \consistency \cloze
    & en 
    & \prex{Dinosaurs? Munich is located [MASK]} 
    & 11k 
    & + \\
& {\cmss ParaRel}~\cite{elazar-etal-2021-measuring} 
    & \consistency \cloze
    & en 
    & \prex{Turing was born in/is native to [MASK]} 
    & n.a.\footnote{number of relations 328, prompts per relation 38}  
    & + \\                               
& \cite{pandia-ettinger-2021-sorting}
    & \consistency \cloze 
    & en
    & \prex{Sebastian lives in France. The capital of Sebastian’s country is [MASK].} 
    & 40k
    & + \\  
& {\cmss SituatedQA (context/answer)} \newline \cite{zhang-choi-2021-situatedqa} 
    & \contextk \questions  
    & en  
    & \prex{Who made the most 3 point shots in the NBA?} 
    & 18k 
    & + \\ 
& \cite{heinzerling-inui-2021-language} 
    & \kb \cloze
    & en 
    & \prex{Turing was born in [MASK]} 
    & 15M 
    & + \\                                           
& \cite{podkorytov-etal-2021-limitations} 
    & \mc \cloze 
    & en 
    & \prex{Tomatoes are a [MASK].} 
    & 0.1k   
    & - \\
& {\cmss mParaRel}~\cite{fierro-sogaard-2022-factual} 
    & \consistency \cloze 
    & multi 
    & \prex{Turing is from/was born in [MASK]} 
    & n.a.\footnote{number of relations 343, prompts per relation 37.13 (avg. over languages)}   
    & + \\
& {\cmss TEMPLAMA}~\cite{dhingra-etal-2022-time} 
    & \contextk  \cloze
    & en  
    & \prex{[2012] Cristiano Ronaldo plays for [MASK]} 
    & 50k 
    & + \\                                                                                       
& \cite{singhania-etal-2022-knowledge} 
    & \kb \cloze 
    & en 
    & \prex{France shares a land border with [MASK]} 
    & 2k 
    & + \\
& \cite{jang-etal-2022-towards} 
    & \ku \cloze
    & en 
    & \prex{[MASK] is the prime minister of England} 
    & 30k 
    & + \\        
& \CR{ \cmss{TemporalWiki}~\cite{jang-etal-2022-temporalwiki}}	
    & \ku \ppl
    & en
    & \prex{On 1 December, the Omicron variant...}
    & \textcircled{d} 
    & o \\

& {\cmss zsRE}~\cite{lee-etal-2022-plug} 
    & \ku \questions 
    & en  
    & \prex{Who is the most paid player in EPL?} 
    & 168k 
    & + \\
& CounterFact~\cite{meng-etal-2022-locating}
   & \ku \cloze
   & en
   & \prex{Turing's mother tongue is <old,new>} 
   & 22k
   & + \\
   
& {\cmss ECBD}~\cite{onoe-etal-2022-entity} 
    & \ue \cloze 
    & en  
    & \prex{[mRNA vaccines] do not affect [MASK].} 
    & 35k 
    & + \\

& \CR{\cite{hase-etal-2023-methods}}
    & \ku \cloze
    & en 
    & \prex{Mary Lowe Good has relation ‘winner of’ to [MASK]} 
    & 170k 
    & + \\

& \CR{ \cmss{CounterFact+}\newline\cite{hoelscher-obermaier-etal-2023-detecting} }
    & \ku \cloze 
    & en 
    & \prex{The mother tongue of Danielle Darrieux is English. The native language of Montesquieu is [MASK]} 
    &  \textcircled{d} 
    & o \\
& \CR{\cmss{DynamicTempLAMA}\newline \cite{margatina-etal-2023-dynamic} }
    & \ku \cloze 
    & en 
    & \prex{The surname of the Prime Minister of the UK is [MASK]}
    & \textcircled{d}
    & + \\ 		
    
\bottomrule
\end{tabular}
\caption{Datasets for factual knowledge probing. 
Probed knowledge: 
 general knowledge \gk, domain-specific knowledge \dk,   context-dependent knowledge \contextk, PLMs sensitivity to paraphrases, negation or mispriming \consistency, related to PLMs as KBs \kb{} , knowledge updating \ku ,  misconceptions \mc{} and  unseen entities \ue. NLP task: cloze prompts \cloze, question answering \questions, classification \cls, natural language generation \nlg, \CR{and perplexity} \ppl. Showing languages, example, and number of instances in the dataset (rounded). Data access: accessible without effort (+), accessible with some effort (o), not accessible (-). \CR{\textcircled{d} refers to dynamic datasets, whose number of instances is changeable over time.} We only include references to papers, in which the datasets are used for factual knowledge probing. References to papers introducing the datasets are added in Table~\ref{app:datasets} in the appendix. 
}
 
 \label{tab:datasets}
\end{minipage}
\end{table*}

We found a variety of datasets (\CR{44} in our corpus) that have been proposed or used for probing factual knowledge in PLMs: \CR{18} datasets for probing general knowledge, 8 for domain-specific knowledge and \CR{18} datasets that target other aspects, e.g, consistency of PLMs (cf. Table~\ref{tab:datasets}).

Datasets for \textbf{general knowledge} probing are used to quantify generic factual knowledge in PLMs with the most prominent being {\cmss LAMA}~\cite{petroni-etal-2019-language}. {\cmss WIKI-UNI}~\cite{cao-etal-2021-knowledgeable} is similar to {\cmss LAMA}, but with a uniform distribution of object entities. {\cmss LAMA-UHN}~\cite{poerner-etal-2020-e} is a subset of LAMA without easy-to-guess examples. \CR{{\cmss{DLAMA}}~\cite{keleg-magdy-2023-dlama} targets culturally diverse facts.} While \CR{16} datasets are solely English, there are \CR{three} multilingual datasets ({\cmss mLAMA}~\cite{kassner-etal-2021-multilingual}, {\cmss X-FACTR}~\cite{jiang-etal-2020-x} and \CR{\cmss{ DLAMA}~\cite{keleg-magdy-2023-dlama})}. {\cmss IndicGLUE}~\cite{kakwani-etal-2020-indicnlpsuite} contains 11 Indic languages. Most datasets consist of cloze prompts, while QA datasets ({\cmss WebQuestions}~\cite{berant-etal-2013-semantic},  {\cmss TriviaQA}~\cite{joshi-etal-2017-triviaqa}, {\cmss NQ}~\cite{kwiatkowski-etal-2019-natural}), \CR{\cmss{PopQA} and \cmss{EntityQuestions}~\cite{mallen-etal-2023-trust}} are also used to quantify factual knowledge~\cite{roberts-etal-2020-much}. \citet{wang-etal-2021-generative} adapt SQuAD~\cite{rajpurkar-etal-2018-know} for closed-book question answering.


6 out of 8 datasets used for probing \textbf{domain-specific} knowledge target the biomedical domain (e.g.,  {\cmss MedQA}~\cite{jin-etal-2021-disease},  {\cmss BioLAMA}~\cite{sung-etal-2021-language} and MedLAMA~\cite{meng-etal-2022-rewire}). The multilingual dataset {\cmss EXAMS}~\cite{hardalov-etal-2020-exams} focuses on scientific QA, whereas  {\cmss LEFT}~\cite{ciosici-etal-2021-perhaps} contains questions from humanities and social sciences. 

The community has constructed further datasets to investigate \textbf{other aspects} of using PLMs as knowledge bases. {\cmss PARAREL}~\cite{elazar-etal-2021-measuring} and its multilingual counterpart {\cmss mPARAREL}~\cite{fierro-sogaard-2022-factual} target the sensitivity of PLMs to paraphrases.  {\cmss Negated/Misprimed LAMA}~\cite{kassner-schutze-2020-negated} focuses on how negation/mispriming affects fact retrieval from PLMs, whereas~\citet{pandia-ettinger-2021-sorting} target the effect of distractors. Updating knowledge in PLMs is considered by~\citet{jang-etal-2022-temporalwiki, jang-etal-2022-towards, lee-etal-2022-plug, meng-etal-2022-locating, hase-etal-2023-methods, hoelscher-obermaier-etal-2023-detecting, margatina-etal-2023-dynamic}.  {\cmss TEMPLAMA}~\cite{dhingra-etal-2022-time} is concerned with time-dependent facts retrieval, whereas  {\cmss SituatedQA}~\cite{zhang-choi-2021-situatedqa} considers both, temporal and geographical contexts. \citet{heinzerling-inui-2021-language} use a large dataset to evaluate the knowledge storing and retrieval capabilities of PLMs, and hence their use as KBs. \citet{singhania-etal-2022-knowledge} challenge the community to build a KB from PLMs, and provide a dataset to facilitate fact retrieval. 


\section{Insights about Knowledge Retention and Prompt Optimization}
\label{sec:insights}

Two further aspects emerged from the surveyed papers: i) factors affecting knowledge retention, and ii) whether prompts should be optimized. 
\subsection{Factors Affecting Knowledge Retention}
\label{ssec:retention}
PLMs are diverse with respect to their architectures, pre-training objectives and their pre-training data. A compelling question is: how do all these factors affect knowledge retention in PLMs?

Large language models are known to perform generally better and hold more knowledge~\cite{brown-etal-2020-language, roberts-etal-2020-much}. However, the model's architecture and pre-training objectives are more decisive for knowledge retention than its size~\cite{li-etal-2022-eliteplm}. For example, pre-training with the Salient Span Masking objective~\cite{guu-etal-2020-realm} helps PLMs to absorb more facts~\cite{roberts-etal-2020-much, cole-etal-2023-salient}. Similarly, \citet{xiong-etal-2020-pretrained} demonstrate that training the model to predict if the original entities in the text have been replaced with other entities is beneficial for fact retrieval. More generally, \citet{ye-etal-2021-influence} conclude that a masking strategy matching the downstream task, positively affects the performance on that task.

A larger pre-training corpus with an encoder-only model~\cite{liu-etal-2020-roberta} leads to higher knowledge retention~\cite{zhang-etal-2021-need}, but with an encoder-decoder model~\cite{lewis-etal-2020-bart}, a larger corpus negatively affects knowledge retention~\citet{wang-etal-2021-generative}. \CR{Recency~\cite{chiang-etal-2020-pretrained} and frequency~\cite{kandpal-etal-2023-large}, i.e., \emph{when} and \emph{how often} the data is observed at training, are also essential for knowledge retention.}


Larger models and more pre-training data can improve knowledge retention if combined with the right choices for architecture and pre-training objective(s). \CR{However, scaling might not be sufficient~\cite{kandpal-etal-2023-large}. Even though many works propose new architectures and pre-training objectives to increase factual knowledge retention in PLMs and their robustness to prompts~\cite{fevry-etal-2020-entities, hosseini-etal-2021-understanding, sadeq-etal-2022-informask, whitehouse-etal-2022-entitycs, min-etal-2023-nonparametric, zhong-etal-2023-self}, this is a promising future work direction, as there is more room for improvement. }

\subsection{Should Prompts be Optimized?}
\label{ssec:optimization}
Prompt Optimizing leads to better probing performance~\cite{jiang-etal-2020-know, shin-etal-2020-autoprompt, kumar-talukdar-2021-reordering, newman-etal-2022-padapters, zhang-etal-2022-promptgen} . 
However, it remains unclear whether this improvement is due to optimized prompts leaking new knowledge into the probed PLMs.

Optimized prompts can be mere paraphrases of manually created prompts~\cite{bouraoui-etal-2020-inducing, jiang-etal-2020-know}. These paraphrases might be better fact retrievers because of their similarity to the pre-training corpus~\cite{cao-etal-2022-prompt}. Other prompt optimization methods find better prompts in discrete or continuous spaces~\cite{shin-etal-2020-autoprompt, zhong-etal-2021-factual}. These prompts are largely uninterpretable, and can even retrieve facts from randomly initialized PLMs~\cite{zhong-etal-2021-factual, ishibashi-etal-2023-evaluating}.  


Performance improvements for optimized prompts can be attributed either to prompts becoming more similar to the pre-training data or overfitting the facts distribution. Evaluation should take the pre-training corpora and the facts distribution in the probing dataset into account~\cite{cao-etal-2021-knowledgeable,cao-etal-2022-prompt}. \CR{Future work should consider adapting prompt optimization methods to produce more interpretable prompts. This would keep the performance gains, and increase the trustworthiness of optimized prompts.}

\section{Obstacles to Adopting PLMs as KBs}
\label{sec:challenges}

\paragraph{Consistency.} A challenge to relying on PLMs as knowledge bases is their sensitivity to the input queries~\cite{fierro-sogaard-2022-factual}. PLMs rely on shallow surface features and lexical correlations~\cite{kassner-schutze-2020-negated, misra-etal-2020-exploring, poerner-etal-2020-e, rogers-etal-2020-primer, li-etal-2022-pre}, which explains their high sensitivity to the way queries are formulated. 
Current solutions~\cite{elazar-etal-2021-measuring, newman-etal-2022-padapters} train PLMs to be robust to variations in inputs, but further improvements are needed to make PLMs reliable knowledge bases. \CR{PLMs are known to be highly sensitive to prompts, especially in languages other than English~\cite{fierro-sogaard-2022-factual}, where less resources are available. Making PLMs more robust to prompts in non-English languages is a promising future work direction.}


\paragraph{Interpretability.} Identifying where facts are stored and how they are retrieved is essential to adopt PLMs as trustworthy knowledge sources.  
Several approaches locate knowledge in PLMs~\cite{singh-etal-2020-bertnesia, podkorytov-etal-2021-limitations, alkhaldi-etal-2022-peek, dai-etal-2022-knowledge, meng-etal-2022-locating}, with different conclusions depending on the architecture (e.g., knowledge is located in the middle layers of GPT-like models~\cite{meng-etal-2022-locating}, or in the upper layers in BERT-like models~\cite{dai-etal-2022-knowledge}). Another line of work focuses on the data aspect, showing the dependence of PLMs on word co-occurrences and positionally close words~\cite{li-etal-2022-pre}, or tracing back predictions to training data~\cite{akyurek-etal-2022-towards, park-etal-2023-trak}. Knowing how PLMs retrieve facts remains challenging, but necessary to make PLMs transparent fact retrievers. \CR{The introduction of a fact tracing benchmark~\cite{akyurek-etal-2022-towards} opens the door for works in this direction.}

\paragraph{Updating Knowledge.} PLMs come with a fixed set of pre-trained parameters that encode knowledge about the world. As time passes, this knowledge becomes partially outdated. Hence, editing existing knowledge in PLMs and augmenting them with new knowledge is crucial for their use as knowledge bases~\cite{zini-etal-2022-explainability}.

One line of research locates the modules responsible for factual predictions and modifies these to update the corresponding facts~\cite{dai-etal-2022-knowledge, de-cao-etal-2021-editing, meng-etal-2022-locating}. Other lines of research keep the original PLM unchanged, but augment it with additional parameters to induce the desired changes~\cite{wang-etal-2021-k, lee-etal-2022-plug}, or encode facts with time stamps in PLMs to make them ``time-aware''~\cite{dhingra-etal-2022-time}.

\CR{When updating facts in PLMs, it is crucial that only the targeted facts are affected and that these facts are retrievable using different paraphrases~\cite{de-cao-etal-2021-editing, hase-etal-2023-methods}. However, current methods for facts editing~\cite{meng-etal-2022-locating, meng-etal-2023-massediting} still do not fulfill these requirements~\cite{hoelscher-obermaier-etal-2023-detecting}. Methods that introduce additional parameters should be made more scalable~\cite{jang-etal-2022-towards}.}

\CR{
\section{Related Work}
\label{sec:relatedwork}
\citet{alkhamissi-etal-2022-review} elaborate requirements for PLMs as knowledge bases and review recent literature w.r.t. those requirements. These requirements are widely known (e.g., consistency~\cite{petroni-etal-2019-language} and updating knowledge~\cite{de-cao-etal-2021-editing}). Our analysis leads to similar general observations (cf. Section~\ref{sec:challenges}), and additionally reviews more recent solutions to these obstacles. \citet{cao-etal-2023-life} cover probing PLMs as part of the knowledge cycle in PLMs, but do not address factual knowledge probing at the same level of detail as we do. ~\citet{liu-etal-2023-prompt-ACM} survey prompting methods in detail. However, they cover only a part of factual knowledge probing methods. \citet{safavi-koutra-2021-relational} survey how PLMs acquire relational knowledge, organizing knowledge representations strategies in PLMs based on different levels of KBs supervision. 
We provide a novel categorization scheme and conduct a systematic analysis of methods for factual knowledge probing that goes beyond all existing surveys.
We additionally provide a categorization of factual probing datasets. Furthermore, we discuss recent findings on knowledge retention, the use of optimized prompts, and challenges with corresponding recent solutions to adopting PLMs as KBs, shedding light on several future work directions. In contrast to other work, we employed a systematic approach to curate and analyze relevant literature to a comprehensive and unbiased representation of existing work.

}

\CR{
\section{Discussion and Future Work}
\label{sec:discussion}
Factual probing methods are developed to extract as many facts as possible from the new smart pools of knowledge, namely PLMs. This gives us an estimate about how much PLMs have learned from pre-training, and help us to assess their suitability for use cases such as PLMs-as-KBs. Improving probing methods should go hand-in-hand with advances in PLMs themselves, to help us better assess and make use of PLMs. Our analysis (cf. Section~\ref{sec:probing}) shows that current probing methods focus mostly on one the the three dimensions we use in our categorization (inputs, PLMs, outputs). Introducing adaptations across two or more of these dimensions (e.g., optimizing inputs while also debiasing outputs) might lead to further improvements with respect to factual knowledge retrieval. 

Besides improving probing methods, it is also essential to pay attention to the benchmark datasets. Some probing datasets are shown to be biased towards certain entities~\cite{cao-etal-2021-knowledgeable}. Constructing unbiased probing datasets is crucial to have unbiased estimates of factual knowledge in PLMs. At the same time, developing comprehensive datasets which correspond to the capacity of the recently published large PLMs, e.g., \cite{openai-2023-gpt4, penedo-etal-2023-falcon, touvron-etal-2023-llama}, is an important future work direction.

We also believe that it is necessary for current evaluation schemes to not be limited to counting how often PLMs answer correctly. Instead, we call for a comprehensive evaluation that includes further important factors such as the number and frequency of the answers in the pre-training corpus, creation period of the pre-training corpus, model size, and the number of training epochs.

}


\clearpage\newpage
\section{Limitations}
\label{sec:limitations}

For our corpus construction we relied on all the publications that cited~\cite{petroni-etal-2019-language}. Although this represents the first work that sparked the community's interest in the factual knowledge present in PLMs and their use as KBs, there might be parallel works or works that go into the same direction but do not directly cite~\citet{petroni-etal-2019-language}'s work, which are not included in our corpus. Additionally, we relied on the venue information provided by Semantic Scholar's API to filter out irrelevant publications. These information are not always accurate and might have affected our initial corpus. 



In this work, we focused on works that revolve around factual knowledge, and excluded works that focus on other types of knowledge (e.g., linguistic knowledge and commonsense knowledge). However, there are methods that are used for other types of knowledge that could also be applied to factual knowledge and vice versa. We consciously excluded works that focused on other types of knowledge, but this does not mean that such methods are not applicable to factual knowledge probing. 

\section*{Acknowledgements}
We thank Jan Trienes, and the three anonymous reviewers for their insightful comments on this work.  
\bibliography{anthology,custom}

\begin{thebibliography}{119}
\expandafter\ifx\csname natexlab\endcsname\relax\def\natexlab#1{#1}\fi

\bibitem[{Abaho et~al.(2022)Abaho, Bollegala, Williamson, and
  Dodd}]{abaho-etal-2022-position}
Micheal Abaho, Danushka Bollegala, Paula Williamson, and Susanna Dodd. 2022.
\newblock \href {https://doi.org/10.18653/v1/2022.bionlp-1.3} {Position-based
  prompting for health outcome generation}.
\newblock In \emph{Proceedings of the 21st Workshop on Biomedical Language
  Processing}, pages 26--36, Dublin, Ireland. Association for Computational
  Linguistics.

\bibitem[{Akyurek et~al.(2022)Akyurek, Bolukbasi, Liu, Xiong, Tenney, Andreas,
  and Guu}]{akyurek-etal-2022-towards}
Ekin Akyurek, Tolga Bolukbasi, Frederick Liu, Binbin Xiong, Ian Tenney, Jacob
  Andreas, and Kelvin Guu. 2022.
\newblock \href {https://doi.org/10.18653/v1/2022.findings-emnlp.180} {Towards
  tracing knowledge in language models back to the training data}.
\newblock In \emph{Findings of the Association for Computational Linguistics:
  EMNLP 2022}, pages 2429--2446, Abu Dhabi, United Arab Emirates. Association
  for Computational Linguistics.

\bibitem[{Alghanmi et~al.(2021)Alghanmi, Espinosa~Anke, and
  Schockaert}]{alghanmi-etal-2021-probing}
Israa Alghanmi, Luis Espinosa~Anke, and Steven Schockaert. 2021.
\newblock \href {https://doi.org/10.18653/v1/2021.findings-acl.266} {Probing
  pre-trained language models for disease knowledge}.
\newblock In \emph{Findings of the Association for Computational Linguistics:
  ACL-IJCNLP 2021}, pages 3023--3033, Online. Association for Computational
  Linguistics.

\bibitem[{Alkhaldi et~al.(2022)Alkhaldi, Chu, and
  Kurohashi}]{alkhaldi-etal-2022-peek}
Tareq Alkhaldi, Chenhui Chu, and Sadao Kurohashi. 2022.
\newblock A peek into the memory of {T5}: Investigating the factual knowledge
  memory in a closed-book qa setting and finding responsible parts.
\newblock \emph{Journal of Natural Language Processing}, 29(3):762--784.

\bibitem[{AlKhamissi et~al.(2022)AlKhamissi, Li, Celikyilmaz, Diab, and
  Ghazvininejad}]{alkhamissi-etal-2022-review}
Badr AlKhamissi, Millicent Li, Asli Celikyilmaz, Mona Diab, and Marjan
  Ghazvininejad. 2022.
\newblock A review on language models as knowledge bases.
\newblock \emph{arXiv preprint arXiv:2204.06031}.

\bibitem[{Berant et~al.(2013)Berant, Chou, Frostig, and
  Liang}]{berant-etal-2013-semantic}
Jonathan Berant, Andrew Chou, Roy Frostig, and Percy Liang. 2013.
\newblock \href {https://aclanthology.org/D13-1160} {Semantic parsing on
  {F}reebase from question-answer pairs}.
\newblock In \emph{Proceedings of the 2013 Conference on Empirical Methods in
  Natural Language Processing}, pages 1533--1544, Seattle, Washington, USA.
  Association for Computational Linguistics.

\bibitem[{Bouraoui et~al.(2020)Bouraoui, Camacho-Collados, and
  Schockaert}]{bouraoui-etal-2020-inducing}
Zied Bouraoui, Jose Camacho-Collados, and Steven Schockaert. 2020.
\newblock Inducing relational knowledge from {{BERT}}.
\newblock In \emph{Proceedings of the AAAI Conference on Artificial
  Intelligence}, volume~34, pages 7456--7463.

\bibitem[{Brown et~al.(2020)Brown, Mann, Ryder, Subbiah, Kaplan, Dhariwal,
  Neelakantan, Shyam, Sastry, Askell et~al.}]{brown-etal-2020-language}
Tom Brown, Benjamin Mann, Nick Ryder, Melanie Subbiah, Jared~D Kaplan, Prafulla
  Dhariwal, Arvind Neelakantan, Pranav Shyam, Girish Sastry, Amanda Askell,
  et~al. 2020.
\newblock Language models are few-shot learners.
\newblock \emph{Advances in neural information processing systems},
  33:1877--1901.

\bibitem[{Cao et~al.(2022)Cao, Lin, Han, Liu, and Sun}]{cao-etal-2022-prompt}
Boxi Cao, Hongyu Lin, Xianpei Han, Fangchao Liu, and Le~Sun. 2022.
\newblock \href {https://doi.org/10.18653/v1/2022.acl-long.398} {Can prompt
  probe pretrained language models? understanding the invisible risks from a
  causal view}.
\newblock In \emph{Proceedings of the 60th Annual Meeting of the Association
  for Computational Linguistics (Volume 1: Long Papers)}, pages 5796--5808,
  Dublin, Ireland. Association for Computational Linguistics.

\bibitem[{Cao et~al.(2023)Cao, Lin, Han, and Sun}]{cao-etal-2023-life}
Boxi Cao, Hongyu Lin, Xianpei Han, and Le~Sun. 2023.
\newblock The life cycle of knowledge in big language models: A survey.
\newblock \emph{arXiv e-prints}, pages arXiv--2303.

\bibitem[{Cao et~al.(2021)Cao, Lin, Han, Sun, Yan, Liao, Xue, and
  Xu}]{cao-etal-2021-knowledgeable}
Boxi Cao, Hongyu Lin, Xianpei Han, Le~Sun, Lingyong Yan, Meng Liao, Tong Xue,
  and Jin Xu. 2021.
\newblock \href {https://doi.org/10.18653/v1/2021.acl-long.146} {Knowledgeable
  or educated guess? revisiting language models as knowledge bases}.
\newblock In \emph{Proceedings of the 59th Annual Meeting of the Association
  for Computational Linguistics and the 11th International Joint Conference on
  Natural Language Processing (Volume 1: Long Papers)}, pages 1860--1874,
  Online. Association for Computational Linguistics.

\bibitem[{Chen et~al.(2022)Chen, Zhong, Zha, Karypis, and
  He}]{chen-etal-2022-meta}
Yanda Chen, Ruiqi Zhong, Sheng Zha, George Karypis, and He~He. 2022.
\newblock \href {https://doi.org/10.18653/v1/2022.acl-long.53} {Meta-learning
  via language model in-context tuning}.
\newblock In \emph{Proceedings of the 60th Annual Meeting of the Association
  for Computational Linguistics (Volume 1: Long Papers)}, pages 719--730,
  Dublin, Ireland. Association for Computational Linguistics.

\bibitem[{Chiang et~al.(2020)Chiang, Huang, and
  Lee}]{chiang-etal-2020-pretrained}
Cheng-Han Chiang, Sung-Feng Huang, and Hung-yi Lee. 2020.
\newblock \href {https://doi.org/10.18653/v1/2020.emnlp-main.553} {{P}retrained
  language model embryology: {T}he birth of {ALBERT}}.
\newblock In \emph{Proceedings of the 2020 Conference on Empirical Methods in
  Natural Language Processing (EMNLP)}, pages 6813--6828, Online. Association
  for Computational Linguistics.

\bibitem[{Chowdhery et~al.(2022)Chowdhery, Narang, Devlin, Bosma, Mishra,
  Roberts, Barham, Chung, Sutton, Gehrmann et~al.}]{chowdhery-etal-2022-palm}
Aakanksha Chowdhery, Sharan Narang, Jacob Devlin, Maarten Bosma, Gaurav Mishra,
  Adam Roberts, Paul Barham, Hyung~Won Chung, Charles Sutton, Sebastian
  Gehrmann, et~al. 2022.
\newblock Pa{LM}: Scaling language modeling with pathways.
\newblock \emph{arXiv preprint arXiv:2204.02311}.

\bibitem[{Ciosici et~al.(2021)Ciosici, Cecil, Lee, Hedges, Freedman, and
  Weischedel}]{ciosici-etal-2021-perhaps}
Manuel Ciosici, Joe Cecil, Dong-Ho Lee, Alex Hedges, Marjorie Freedman, and
  Ralph Weischedel. 2021.
\newblock \href {https://doi.org/10.18653/v1/2021.emnlp-main.493} {Perhaps
  {PTLM}s should go to school {--} a task to assess open book and closed book
  {QA}}.
\newblock In \emph{Proceedings of the 2021 Conference on Empirical Methods in
  Natural Language Processing}, pages 6104--6111, Online and Punta Cana,
  Dominican Republic. Association for Computational Linguistics.

\bibitem[{Cole et~al.(2023)Cole, Chaudhary, Dhingra, and
  Talukdar}]{cole-etal-2023-salient}
Jeremy~R. Cole, Aditi Chaudhary, Bhuwan Dhingra, and Partha Talukdar. 2023.
\newblock \href {https://doi.org/10.18653/v1/2023.eacl-main.222} {Salient span
  masking for temporal understanding}.
\newblock In \emph{Proceedings of the 17th Conference of the European Chapter
  of the Association for Computational Linguistics}, pages 3052--3060,
  Dubrovnik, Croatia. Association for Computational Linguistics.

\bibitem[{Dai et~al.(2022)Dai, Dong, Hao, Sui, Chang, and
  Wei}]{dai-etal-2022-knowledge}
Damai Dai, Li~Dong, Yaru Hao, Zhifang Sui, Baobao Chang, and Furu Wei. 2022.
\newblock \href {https://doi.org/10.18653/v1/2022.acl-long.581} {Knowledge
  neurons in pretrained transformers}.
\newblock In \emph{Proceedings of the 60th Annual Meeting of the Association
  for Computational Linguistics (Volume 1: Long Papers)}, pages 8493--8502,
  Dublin, Ireland. Association for Computational Linguistics.

\bibitem[{Davison et~al.(2019)Davison, Feldman, and
  Rush}]{davison-etal-2019-commonsense}
Joe Davison, Joshua Feldman, and Alexander Rush. 2019.
\newblock \href {https://doi.org/10.18653/v1/D19-1109} {Commonsense knowledge
  mining from pretrained models}.
\newblock In \emph{Proceedings of the 2019 Conference on Empirical Methods in
  Natural Language Processing and the 9th International Joint Conference on
  Natural Language Processing (EMNLP-IJCNLP)}, pages 1173--1178, Hong Kong,
  China. Association for Computational Linguistics.

\bibitem[{De~Cao et~al.(2021)De~Cao, Aziz, and
  Titov}]{de-cao-etal-2021-editing}
Nicola De~Cao, Wilker Aziz, and Ivan Titov. 2021.
\newblock \href {https://doi.org/10.18653/v1/2021.emnlp-main.522} {Editing
  factual knowledge in language models}.
\newblock In \emph{Proceedings of the 2021 Conference on Empirical Methods in
  Natural Language Processing}, pages 6491--6506, Online and Punta Cana,
  Dominican Republic. Association for Computational Linguistics.

\bibitem[{Devlin et~al.(2019)Devlin, Chang, Lee, and
  Toutanova}]{devlin-etal-2019-bert}
Jacob Devlin, Ming-Wei Chang, Kenton Lee, and Kristina Toutanova. 2019.
\newblock \href {https://doi.org/10.18653/v1/N19-1423} {{BERT}: Pre-training of
  deep bidirectional transformers for language understanding}.
\newblock In \emph{Proceedings of the 2019 Conference of the North {A}merican
  Chapter of the Association for Computational Linguistics: Human Language
  Technologies, Volume 1 (Long and Short Papers)}, pages 4171--4186,
  Minneapolis, Minnesota. Association for Computational Linguistics.

\bibitem[{Dhingra et~al.(2022)Dhingra, Cole, Eisenschlos, Gillick, Eisenstein,
  and Cohen}]{dhingra-etal-2022-time}
Bhuwan Dhingra, Jeremy~R. Cole, Julian~Martin Eisenschlos, Daniel Gillick,
  Jacob Eisenstein, and William~W. Cohen. 2022.
\newblock \href {https://doi.org/10.1162/tacl_a_00459} {Time-aware language
  models as temporal knowledge bases}.
\newblock \emph{Transactions of the Association for Computational Linguistics},
  10:257--273.

\bibitem[{Dong et~al.(2022)Dong, Dai, Song, Xu, Sui, and
  Li}]{dong-etal-2022-calibrating}
Qingxiu Dong, Damai Dai, Yifan Song, Jingjing Xu, Zhifang Sui, and Lei Li.
  2022.
\newblock \href {https://doi.org/10.18653/v1/2022.findings-emnlp.438}
  {Calibrating factual knowledge in pretrained language models}.
\newblock In \emph{Findings of the Association for Computational Linguistics:
  EMNLP 2022}, pages 5937--5947, Abu Dhabi, United Arab Emirates. Association
  for Computational Linguistics.

\bibitem[{Dufter et~al.(2021)Dufter, Kassner, and
  Sch{\"u}tze}]{dufter-etal-2021-static}
Philipp Dufter, Nora Kassner, and Hinrich Sch{\"u}tze. 2021.
\newblock \href {https://doi.org/10.18653/v1/2021.naacl-main.186} {Static
  embeddings as efficient knowledge bases?}
\newblock In \emph{Proceedings of the 2021 Conference of the North American
  Chapter of the Association for Computational Linguistics: Human Language
  Technologies}, pages 2353--2363, Online. Association for Computational
  Linguistics.

\bibitem[{Elazar et~al.(2021)Elazar, Kassner, Ravfogel, Ravichander, Hovy,
  Sch{\"u}tze, and Goldberg}]{elazar-etal-2021-measuring}
Yanai Elazar, Nora Kassner, Shauli Ravfogel, Abhilasha Ravichander, Eduard
  Hovy, Hinrich Sch{\"u}tze, and Yoav Goldberg. 2021.
\newblock \href {https://doi.org/10.1162/tacl_a_00410} {Measuring and improving
  consistency in pretrained language models}.
\newblock \emph{Transactions of the Association for Computational Linguistics},
  9:1012--1031.

\bibitem[{F{\'e}vry et~al.(2020)F{\'e}vry, Baldini~Soares, FitzGerald, Choi,
  and Kwiatkowski}]{fevry-etal-2020-entities}
Thibault F{\'e}vry, Livio Baldini~Soares, Nicholas FitzGerald, Eunsol Choi, and
  Tom Kwiatkowski. 2020.
\newblock \href {https://doi.org/10.18653/v1/2020.emnlp-main.400} {Entities as
  experts: Sparse memory access with entity supervision}.
\newblock In \emph{Proceedings of the 2020 Conference on Empirical Methods in
  Natural Language Processing (EMNLP)}, pages 4937--4951, Online. Association
  for Computational Linguistics.

\bibitem[{Fichtel et~al.(2021)Fichtel, Kalo, and
  Balke}]{fichtel-etal-2021-prompt}
Leandra Fichtel, Jan-Christoph Kalo, and Wolf-Tilo Balke. 2021.
\newblock \href {https://openreview.net/forum?id=o7sMlpr9yBW} {Prompt tuning or
  fine-tuning - investigating relational knowledge in pre-trained language
  models}.
\newblock In \emph{3rd Conference on Automated Knowledge Base Construction}.

\bibitem[{Fierro and S{\o}gaard(2022)}]{fierro-sogaard-2022-factual}
Constanza Fierro and Anders S{\o}gaard. 2022.
\newblock \href {https://doi.org/10.18653/v1/2022.findings-acl.240} {Factual
  consistency of multilingual pretrained language models}.
\newblock In \emph{Findings of the Association for Computational Linguistics:
  ACL 2022}, pages 3046--3052, Dublin, Ireland. Association for Computational
  Linguistics.

\bibitem[{Gladkova et~al.(2016)Gladkova, Drozd, and
  Matsuoka}]{gladkova-etal-2016-analogy}
Anna Gladkova, Aleksandr Drozd, and Satoshi Matsuoka. 2016.
\newblock \href {https://doi.org/10.18653/v1/N16-2002} {Analogy-based detection
  of morphological and semantic relations with word embeddings: what works and
  what doesn{'}t.}
\newblock In \emph{Proceedings of the {NAACL} Student Research Workshop}, pages
  8--15, San Diego, California. Association for Computational Linguistics.

\bibitem[{Guu et~al.(2020)Guu, Lee, Tung, Pasupat, and
  Chang}]{guu-etal-2020-realm}
Kelvin Guu, Kenton Lee, Zora Tung, Panupong Pasupat, and Ming-Wei Chang. 2020.
\newblock {REALM}: Retrieval-augmented language model pre-training.
\newblock In \emph{Proceedings of the 37th International Conference on Machine
  Learning}, ICML'20. JMLR.org.

\bibitem[{Han et~al.(2021)Han, Zhang, Ding, Gu, Liu, Huo, Qiu, Yao, Zhang,
  Zhang, Han, Huang, Jin, Lan, Liu, Liu, Lu, Qiu, Song, Tang, Wen, Yuan, Zhao,
  and Zhu}]{han-etal-2021-pretrained}
Xu~Han, Zhengyan Zhang, Ning Ding, Yuxian Gu, Xiao Liu, Yuqi Huo, Jiezhong Qiu,
  Yuan Yao, Ao~Zhang, Liang Zhang, Wentao Han, Minlie Huang, Qin Jin, Yanyan
  Lan, Yang Liu, Zhiyuan Liu, Zhiwu Lu, Xipeng Qiu, Ruihua Song, Jie Tang,
  Ji-Rong Wen, Jinhui Yuan, Wayne~Xin Zhao, and Jun Zhu. 2021.
\newblock \href {https://doi.org/https://doi.org/10.1016/j.aiopen.2021.08.002}
  {Pre-trained models: Past, present and future}.
\newblock \emph{AI Open}, 2:225--250.

\bibitem[{Hardalov et~al.(2020)Hardalov, Mihaylov, Zlatkova, Dinkov, Koychev,
  and Nakov}]{hardalov-etal-2020-exams}
Momchil Hardalov, Todor Mihaylov, Dimitrina Zlatkova, Yoan Dinkov, Ivan
  Koychev, and Preslav Nakov. 2020.
\newblock \href {https://doi.org/10.18653/v1/2020.emnlp-main.438} {{EXAMS}: A
  multi-subject high school examinations dataset for cross-lingual and
  multilingual question answering}.
\newblock In \emph{Proceedings of the 2020 Conference on Empirical Methods in
  Natural Language Processing (EMNLP)}, pages 5427--5444, Online. Association
  for Computational Linguistics.

\bibitem[{Hase et~al.(2023)Hase, Diab, Celikyilmaz, Li, Kozareva, Stoyanov,
  Bansal, and Iyer}]{hase-etal-2023-methods}
Peter Hase, Mona Diab, Asli Celikyilmaz, Xian Li, Zornitsa Kozareva, Veselin
  Stoyanov, Mohit Bansal, and Srinivasan Iyer. 2023.
\newblock \href {https://doi.org/10.18653/v1/2023.eacl-main.199} {Methods for
  measuring, updating, and visualizing factual beliefs in language models}.
\newblock In \emph{Proceedings of the 17th Conference of the European Chapter
  of the Association for Computational Linguistics}, pages 2714--2731,
  Dubrovnik, Croatia. Association for Computational Linguistics.

\bibitem[{Haviv et~al.(2021)Haviv, Berant, and
  Globerson}]{haviv-etal-2021-bertese}
Adi Haviv, Jonathan Berant, and Amir Globerson. 2021.
\newblock \href {https://doi.org/10.18653/v1/2021.eacl-main.316} {{BERT}ese:
  Learning to speak to {BERT}}.
\newblock In \emph{Proceedings of the 16th Conference of the European Chapter
  of the Association for Computational Linguistics: Main Volume}, pages
  3618--3623, Online. Association for Computational Linguistics.

\bibitem[{Heinzerling and Inui(2021)}]{heinzerling-inui-2021-language}
Benjamin Heinzerling and Kentaro Inui. 2021.
\newblock \href {https://doi.org/10.18653/v1/2021.eacl-main.153} {Language
  models as knowledge bases: On entity representations, storage capacity, and
  paraphrased queries}.
\newblock In \emph{Proceedings of the 16th Conference of the European Chapter
  of the Association for Computational Linguistics: Main Volume}, pages
  1772--1791, Online. Association for Computational Linguistics.

\bibitem[{Hoelscher-Obermaier et~al.(2023)Hoelscher-Obermaier, Persson, Kran,
  Konstas, and Barez}]{hoelscher-obermaier-etal-2023-detecting}
Jason Hoelscher-Obermaier, Julia Persson, Esben Kran, Ioannis Konstas, and Fazl
  Barez. 2023.
\newblock \href {https://doi.org/10.18653/v1/2023.findings-acl.733} {Detecting
  edit failures in large language models: An improved specificity benchmark}.
\newblock In \emph{Findings of the Association for Computational Linguistics:
  ACL 2023}, pages 11548--11559, Toronto, Canada. Association for Computational
  Linguistics.

\bibitem[{Hosseini et~al.(2021)Hosseini, Reddy, Bahdanau, Hjelm, Sordoni, and
  Courville}]{hosseini-etal-2021-understanding}
Arian Hosseini, Siva Reddy, Dzmitry Bahdanau, R~Devon Hjelm, Alessandro
  Sordoni, and Aaron Courville. 2021.
\newblock \href {https://doi.org/10.18653/v1/2021.naacl-main.102}
  {Understanding by understanding not: Modeling negation in language models}.
\newblock In \emph{Proceedings of the 2021 Conference of the North American
  Chapter of the Association for Computational Linguistics: Human Language
  Technologies}, pages 1301--1312, Online. Association for Computational
  Linguistics.

\bibitem[{Houlsby et~al.(2019)Houlsby, Giurgiu, Jastrzebski, Morrone,
  De~Laroussilhe, Gesmundo, Attariyan, and Gelly}]{houlsby-etal-2019-adapters}
Neil Houlsby, Andrei Giurgiu, Stanislaw Jastrzebski, Bruna Morrone, Quentin
  De~Laroussilhe, Andrea Gesmundo, Mona Attariyan, and Sylvain Gelly. 2019.
\newblock Parameter-efficient transfer learning for {NLP}.
\newblock In \emph{International Conference on Machine Learning}, pages
  2790--2799. PMLR.

\bibitem[{Ishibashi et~al.(2023)Ishibashi, Bollegala, Sudoh, and
  Nakamura}]{ishibashi-etal-2023-evaluating}
Yoichi Ishibashi, Danushka Bollegala, Katsuhito Sudoh, and Satoshi Nakamura.
  2023.
\newblock \href {https://doi.org/10.18653/v1/2023.eacl-main.174} {Evaluating
  the robustness of discrete prompts}.
\newblock In \emph{Proceedings of the 17th Conference of the European Chapter
  of the Association for Computational Linguistics}, pages 2373--2384,
  Dubrovnik, Croatia. Association for Computational Linguistics.

\bibitem[{Jang et~al.(2022{\natexlab{a}})Jang, Ye, Lee, Yang, Shin, Han, Kim,
  and Seo}]{jang-etal-2022-temporalwiki}
Joel Jang, Seonghyeon Ye, Changho Lee, Sohee Yang, Joongbo Shin, Janghoon Han,
  Gyeonghun Kim, and Minjoon Seo. 2022{\natexlab{a}}.
\newblock \href {https://doi.org/10.18653/v1/2022.emnlp-main.418}
  {{T}emporal{W}iki: A lifelong benchmark for training and evaluating
  ever-evolving language models}.
\newblock In \emph{Proceedings of the 2022 Conference on Empirical Methods in
  Natural Language Processing}, pages 6237--6250, Abu Dhabi, United Arab
  Emirates. Association for Computational Linguistics.

\bibitem[{Jang et~al.(2022{\natexlab{b}})Jang, Ye, Yang, Shin, Han, KIM, Choi,
  and Seo}]{jang-etal-2022-towards}
Joel Jang, Seonghyeon Ye, Sohee Yang, Joongbo Shin, Janghoon Han, Gyeonghun
  KIM, Stanley~Jungkyu Choi, and Minjoon Seo. 2022{\natexlab{b}}.
\newblock \href {https://openreview.net/forum?id=vfsRB5MImo9} {Towards
  continual knowledge learning of language models}.
\newblock In \emph{International Conference on Learning Representations}.

\bibitem[{Jawahar et~al.(2019)Jawahar, Sagot, and
  Seddah}]{jawahar-etal-2019-bert}
Ganesh Jawahar, Beno{\^\i}t Sagot, and Djam{\'e} Seddah. 2019.
\newblock \href {https://doi.org/10.18653/v1/P19-1356} {What does {BERT} learn
  about the structure of language?}
\newblock In \emph{Proceedings of the 57th Annual Meeting of the Association
  for Computational Linguistics}, pages 3651--3657, Florence, Italy.
  Association for Computational Linguistics.

\bibitem[{Jiang et~al.(2020{\natexlab{a}})Jiang, Anastasopoulos, Araki, Ding,
  and Neubig}]{jiang-etal-2020-x}
Zhengbao Jiang, Antonios Anastasopoulos, Jun Araki, Haibo Ding, and Graham
  Neubig. 2020{\natexlab{a}}.
\newblock \href {https://doi.org/10.18653/v1/2020.emnlp-main.479} {{X}-{FACTR}:
  Multilingual factual knowledge retrieval from pretrained language models}.
\newblock In \emph{Proceedings of the 2020 Conference on Empirical Methods in
  Natural Language Processing (EMNLP)}, pages 5943--5959, Online. Association
  for Computational Linguistics.

\bibitem[{Jiang et~al.(2020{\natexlab{b}})Jiang, Xu, Araki, and
  Neubig}]{jiang-etal-2020-know}
Zhengbao Jiang, Frank~F. Xu, Jun Araki, and Graham Neubig. 2020{\natexlab{b}}.
\newblock \href {https://doi.org/10.1162/tacl_a_00324} {How can we know what
  language models know?}
\newblock \emph{Transactions of the Association for Computational Linguistics},
  8:423--438.

\bibitem[{Jin et~al.(2021)Jin, Pan, Oufattole, Weng, Fang, and
  Szolovits}]{jin-etal-2021-disease}
Di~Jin, Eileen Pan, Nassim Oufattole, Wei-Hung Weng, Hanyi Fang, and Peter
  Szolovits. 2021.
\newblock What disease does this patient have? a large-scale open domain
  question answering dataset from medical exams.
\newblock \emph{Applied Sciences}, 11(14):6421.

\bibitem[{Joshi et~al.(2017)Joshi, Choi, Weld, and
  Zettlemoyer}]{joshi-etal-2017-triviaqa}
Mandar Joshi, Eunsol Choi, Daniel Weld, and Luke Zettlemoyer. 2017.
\newblock \href {https://doi.org/10.18653/v1/P17-1147} {{T}rivia{QA}: A large
  scale distantly supervised challenge dataset for reading comprehension}.
\newblock In \emph{Proceedings of the 55th Annual Meeting of the Association
  for Computational Linguistics (Volume 1: Long Papers)}, pages 1601--1611,
  Vancouver, Canada. Association for Computational Linguistics.

\bibitem[{Kakwani et~al.(2020)Kakwani, Kunchukuttan, Golla, N.C.,
  Bhattacharyya, Khapra, and Kumar}]{kakwani-etal-2020-indicnlpsuite}
Divyanshu Kakwani, Anoop Kunchukuttan, Satish Golla, Gokul N.C., Avik
  Bhattacharyya, Mitesh~M. Khapra, and Pratyush Kumar. 2020.
\newblock \href {https://doi.org/10.18653/v1/2020.findings-emnlp.445}
  {{I}ndic{NLPS}uite: Monolingual corpora, evaluation benchmarks and
  pre-trained multilingual language models for {I}ndian languages}.
\newblock In \emph{Findings of the Association for Computational Linguistics:
  EMNLP 2020}, pages 4948--4961, Online. Association for Computational
  Linguistics.

\bibitem[{Kalinsky et~al.(2023)Kalinsky, Kushilevitz, Libov, and
  Goldberg}]{kalinsky-etal-2023-simple}
Oren Kalinsky, Guy Kushilevitz, Alexander Libov, and Yoav Goldberg. 2023.
\newblock \href {https://doi.org/10.18653/v1/2023.findings-eacl.179} {Simple
  and effective multi-token completion from masked language models}.
\newblock In \emph{Findings of the Association for Computational Linguistics:
  EACL 2023}, pages 2356--2369, Dubrovnik, Croatia. Association for
  Computational Linguistics.

\bibitem[{Kalo and Fichtel(2022)}]{kalo-fichtel-2022-kamel}
Jan-Christoph Kalo and Leandra Fichtel. 2022.
\newblock {KAMEL}: Knowledge analysis with multitoken entities in language
  models.
\newblock In \emph{Proceedings of the Conference on Automated Knowledge Base
  Construction}.

\bibitem[{Kalyan et~al.(2021)Kalyan, Rajasekharan, and
  Sangeetha}]{kalyan-etal-2021-ammus}
Katikapalli~Subramanyam Kalyan, Ajit Rajasekharan, and Sivanesan Sangeetha.
  2021.
\newblock \href {http://arxiv.org/abs/2108.05542} {{AMMUS} : A survey of
  transformer-based pretrained models in natural language processing}.

\bibitem[{Kandpal et~al.(2023)Kandpal, Deng, Roberts, Wallace, and
  Raffel}]{kandpal-etal-2023-large}
Nikhil Kandpal, Haikang Deng, Adam Roberts, Eric Wallace, and Colin Raffel.
  2023.
\newblock \href {https://proceedings.mlr.press/v202/kandpal23a.html} {Large
  language models struggle to learn long-tail knowledge}.
\newblock In \emph{Proceedings of the 40th International Conference on Machine
  Learning}, volume 202 of \emph{Proceedings of Machine Learning Research},
  pages 15696--15707. PMLR.

\bibitem[{Kassner et~al.(2021)Kassner, Dufter, and
  Sch{\"u}tze}]{kassner-etal-2021-multilingual}
Nora Kassner, Philipp Dufter, and Hinrich Sch{\"u}tze. 2021.
\newblock \href {https://doi.org/10.18653/v1/2021.eacl-main.284} {Multilingual
  {LAMA}: Investigating knowledge in multilingual pretrained language models}.
\newblock In \emph{Proceedings of the 16th Conference of the European Chapter
  of the Association for Computational Linguistics: Main Volume}, pages
  3250--3258, Online. Association for Computational Linguistics.

\bibitem[{Kassner and Sch{\"u}tze(2020)}]{kassner-schutze-2020-negated}
Nora Kassner and Hinrich Sch{\"u}tze. 2020.
\newblock \href {https://doi.org/10.18653/v1/2020.acl-main.698} {Negated and
  misprimed probes for pretrained language models: Birds can talk, but cannot
  fly}.
\newblock In \emph{Proceedings of the 58th Annual Meeting of the Association
  for Computational Linguistics}, pages 7811--7818, Online. Association for
  Computational Linguistics.

\bibitem[{Keleg and Magdy(2023)}]{keleg-magdy-2023-dlama}
Amr Keleg and Walid Magdy. 2023.
\newblock \href {https://doi.org/10.18653/v1/2023.findings-acl.389} {{DLAMA}: A
  framework for curating culturally diverse facts for probing the knowledge of
  pretrained language models}.
\newblock In \emph{Findings of the Association for Computational Linguistics:
  ACL 2023}, pages 6245--6266, Toronto, Canada. Association for Computational
  Linguistics.

\bibitem[{Kumar and Talukdar(2021)}]{kumar-talukdar-2021-reordering}
Sawan Kumar and Partha Talukdar. 2021.
\newblock \href {https://doi.org/10.18653/v1/2021.findings-acl.395} {Reordering
  examples helps during priming-based few-shot learning}.
\newblock In \emph{Findings of the Association for Computational Linguistics:
  ACL-IJCNLP 2021}, pages 4507--4518, Online. Association for Computational
  Linguistics.

\bibitem[{Kwiatkowski et~al.(2019)Kwiatkowski, Palomaki, Redfield, Collins,
  Parikh, Alberti, Epstein, Polosukhin, Devlin, Lee, Toutanova, Jones, Kelcey,
  Chang, Dai, Uszkoreit, Le, and Petrov}]{kwiatkowski-etal-2019-natural}
Tom Kwiatkowski, Jennimaria Palomaki, Olivia Redfield, Michael Collins, Ankur
  Parikh, Chris Alberti, Danielle Epstein, Illia Polosukhin, Jacob Devlin,
  Kenton Lee, Kristina Toutanova, Llion Jones, Matthew Kelcey, Ming-Wei Chang,
  Andrew~M. Dai, Jakob Uszkoreit, Quoc Le, and Slav Petrov. 2019.
\newblock \href {https://doi.org/10.1162/tacl_a_00276} {Natural questions: A
  benchmark for question answering research}.
\newblock \emph{Transactions of the Association for Computational Linguistics},
  7:452--466.

\bibitem[{Lee et~al.(2022)Lee, Han, Hwang, Lee, Park, and
  Lee}]{lee-etal-2022-plug}
Kyungjae Lee, Wookje Han, Seung-won Hwang, Hwaran Lee, Joonsuk Park, and
  Sang-Woo Lee. 2022.
\newblock \href {https://doi.org/10.18653/v1/2022.findings-acl.37}
  {Plug-and-play adaptation for continuously-updated {QA}}.
\newblock In \emph{Findings of the Association for Computational Linguistics:
  ACL 2022}, pages 438--447, Dublin, Ireland. Association for Computational
  Linguistics.

\bibitem[{Levy et~al.(2017)Levy, Seo, Choi, and
  Zettlemoyer}]{levy-etal-2017-zero}
Omer Levy, Minjoon Seo, Eunsol Choi, and Luke Zettlemoyer. 2017.
\newblock \href {https://doi.org/10.18653/v1/K17-1034} {Zero-shot relation
  extraction via reading comprehension}.
\newblock In \emph{Proceedings of the 21st Conference on Computational Natural
  Language Learning ({C}o{NLL} 2017)}, pages 333--342, Vancouver, Canada.
  Association for Computational Linguistics.

\bibitem[{Lewis et~al.(2020)Lewis, Liu, Goyal, Ghazvininejad, Mohamed, Levy,
  Stoyanov, and Zettlemoyer}]{lewis-etal-2020-bart}
Mike Lewis, Yinhan Liu, Naman Goyal, Marjan Ghazvininejad, Abdelrahman Mohamed,
  Omer Levy, Veselin Stoyanov, and Luke Zettlemoyer. 2020.
\newblock \href {https://doi.org/10.18653/v1/2020.acl-main.703} {{BART}:
  Denoising sequence-to-sequence pre-training for natural language generation,
  translation, and comprehension}.
\newblock In \emph{Proceedings of the 58th Annual Meeting of the Association
  for Computational Linguistics}, pages 7871--7880, Online. Association for
  Computational Linguistics.

\bibitem[{Lewis et~al.(2021)Lewis, Stenetorp, and
  Riedel}]{lewis-etal-2021-question}
Patrick Lewis, Pontus Stenetorp, and Sebastian Riedel. 2021.
\newblock \href {https://doi.org/10.18653/v1/2021.eacl-main.86} {Question and
  answer test-train overlap in open-domain question answering datasets}.
\newblock In \emph{Proceedings of the 16th Conference of the European Chapter
  of the Association for Computational Linguistics: Main Volume}, pages
  1000--1008, Online. Association for Computational Linguistics.

\bibitem[{Li et~al.(2022{\natexlab{a}})Li, Tang, Gong, Yang, Yu, Chen, Wang,
  Zhao, and Wen}]{li-etal-2022-eliteplm}
Junyi Li, Tianyi Tang, Zheng Gong, Lixin Yang, Zhuohao Yu, Zhipeng Chen,
  Jingyuan Wang, Xin Zhao, and Ji-Rong Wen. 2022{\natexlab{a}}.
\newblock \href {https://doi.org/10.18653/v1/2022.naacl-main.258}
  {{E}lite{PLM}: An empirical study on general language ability evaluation of
  pretrained language models}.
\newblock In \emph{Proceedings of the 2022 Conference of the North American
  Chapter of the Association for Computational Linguistics: Human Language
  Technologies}, pages 3519--3539, Seattle, United States. Association for
  Computational Linguistics.

\bibitem[{Li et~al.(2022{\natexlab{b}})Li, Li, Shang, Dong, Sun, Liu, Ji,
  Jiang, and Liu}]{li-etal-2022-pre}
Shaobo Li, Xiaoguang Li, Lifeng Shang, Zhenhua Dong, Chengjie Sun, Bingquan
  Liu, Zhenzhou Ji, Xin Jiang, and Qun Liu. 2022{\natexlab{b}}.
\newblock \href {https://doi.org/10.18653/v1/2022.findings-acl.136} {How
  pre-trained language models capture factual knowledge? a causal-inspired
  analysis}.
\newblock In \emph{Findings of the Association for Computational Linguistics:
  ACL 2022}, pages 1720--1732, Dublin, Ireland. Association for Computational
  Linguistics.

\bibitem[{Li et~al.(2022{\natexlab{c}})Li, Che, Wang, Jiang, Xiong, and
  Chaturvedi}]{li-etal-2022-spe}
Yiyuan Li, Tong Che, Yezhen Wang, Zhengbao Jiang, Caiming Xiong, and Snigdha
  Chaturvedi. 2022{\natexlab{c}}.
\newblock \href {https://doi.org/10.18653/v1/2022.emnlp-main.803} {{SPE}:
  Symmetrical prompt enhancement for fact probing}.
\newblock In \emph{Proceedings of the 2022 Conference on Empirical Methods in
  Natural Language Processing}, pages 11689--11698, Abu Dhabi, United Arab
  Emirates. Association for Computational Linguistics.

\bibitem[{Li{\'e}tard et~al.(2021)Li{\'e}tard, Abdou, and
  S{\o}gaard}]{lietard-etal-2021-language}
Bastien Li{\'e}tard, Mostafa Abdou, and Anders S{\o}gaard. 2021.
\newblock \href {https://doi.org/10.18653/v1/2021.blackboxnlp-1.40} {Do
  language models know the way to {R}ome?}
\newblock In \emph{Proceedings of the Fourth BlackboxNLP Workshop on Analyzing
  and Interpreting Neural Networks for NLP}, pages 510--517, Punta Cana,
  Dominican Republic. Association for Computational Linguistics.

\bibitem[{Liu et~al.(2023)Liu, Yuan, Fu, Jiang, Hayashi, and
  Neubig}]{liu-etal-2023-prompt-ACM}
Pengfei Liu, Weizhe Yuan, Jinlan Fu, Zhengbao Jiang, Hiroaki Hayashi, and
  Graham Neubig. 2023.
\newblock \href {https://doi.org/10.1145/3560815} {Pre-train, prompt, and
  predict: A systematic survey of prompting methods in natural language
  processing}.
\newblock \emph{ACM Comput. Surv.}, 55(9).

\bibitem[{Liu et~al.(2020)Liu, Ott, Goyal, Du, Joshi, Chen, Levy, Lewis,
  Zettlemoyer, and Stoyanov}]{liu-etal-2020-roberta}
Yinhan Liu, Myle Ott, Naman Goyal, Jingfei Du, Mandar Joshi, Danqi Chen, Omer
  Levy, Mike Lewis, Luke Zettlemoyer, and Veselin Stoyanov. 2020.
\newblock \href {https://openreview.net/forum?id=SyxS0T4tvS} {Ro{BERT}a: A
  robustly optimized {BERT} pretraining approach}.

\bibitem[{Longpre et~al.(2021)Longpre, Perisetla, Chen, Ramesh, DuBois, and
  Singh}]{longpre-etal-2021-entity}
Shayne Longpre, Kartik Perisetla, Anthony Chen, Nikhil Ramesh, Chris DuBois,
  and Sameer Singh. 2021.
\newblock \href {https://doi.org/10.18653/v1/2021.emnlp-main.565} {Entity-based
  knowledge conflicts in question answering}.
\newblock In \emph{Proceedings of the 2021 Conference on Empirical Methods in
  Natural Language Processing}, pages 7052--7063, Online and Punta Cana,
  Dominican Republic. Association for Computational Linguistics.

\bibitem[{Malkin et~al.(2022)Malkin, Wang, and
  Jojic}]{malkin-etal-2022-coherence}
Nikolay Malkin, Zhen Wang, and Nebojsa Jojic. 2022.
\newblock \href {https://doi.org/10.18653/v1/2022.acl-long.565} {Coherence
  boosting: When your pretrained language model is not paying enough
  attention}.
\newblock In \emph{Proceedings of the 60th Annual Meeting of the Association
  for Computational Linguistics (Volume 1: Long Papers)}, pages 8214--8236,
  Dublin, Ireland. Association for Computational Linguistics.

\bibitem[{Mallen et~al.(2023)Mallen, Asai, Zhong, Das, Khashabi, and
  Hajishirzi}]{mallen-etal-2023-trust}
Alex Mallen, Akari Asai, Victor Zhong, Rajarshi Das, Daniel Khashabi, and
  Hannaneh Hajishirzi. 2023.
\newblock \href {https://doi.org/10.18653/v1/2023.acl-long.546} {When not to
  trust language models: Investigating effectiveness of parametric and
  non-parametric memories}.
\newblock In \emph{Proceedings of the 61st Annual Meeting of the Association
  for Computational Linguistics (Volume 1: Long Papers)}, pages 9802--9822,
  Toronto, Canada. Association for Computational Linguistics.

\bibitem[{Margatina et~al.(2023)Margatina, Wang, Vyas, Anna~John, Benajiba, and
  Ballesteros}]{margatina-etal-2023-dynamic}
Katerina Margatina, Shuai Wang, Yogarshi Vyas, Neha Anna~John, Yassine
  Benajiba, and Miguel Ballesteros. 2023.
\newblock \href {https://doi.org/10.18653/v1/2023.eacl-main.211} {Dynamic
  benchmarking of masked language models on temporal concept drift with
  multiple views}.
\newblock In \emph{Proceedings of the 17th Conference of the European Chapter
  of the Association for Computational Linguistics}, pages 2881--2898,
  Dubrovnik, Croatia. Association for Computational Linguistics.

\bibitem[{Meng et~al.(2022{\natexlab{a}})Meng, Bau, Andonian, and
  Belinkov}]{meng-etal-2022-locating}
Kevin Meng, David Bau, Alex~J Andonian, and Yonatan Belinkov.
  2022{\natexlab{a}}.
\newblock \href {https://openreview.net/forum?id=-h6WAS6eE4} {Locating and
  editing factual associations in {GPT}}.
\newblock In \emph{Advances in Neural Information Processing Systems}.

\bibitem[{Meng et~al.(2023)Meng, Sharma, Andonian, Belinkov, and
  Bau}]{meng-etal-2023-massediting}
Kevin Meng, Arnab~Sen Sharma, Alex~J Andonian, Yonatan Belinkov, and David Bau.
  2023.
\newblock \href {https://openreview.net/forum?id=MkbcAHIYgyS} {Mass-editing
  memory in a transformer}.
\newblock In \emph{The Eleventh International Conference on Learning
  Representations}.

\bibitem[{Meng et~al.(2022{\natexlab{b}})Meng, Liu, Shareghi, Su, Collins, and
  Collier}]{meng-etal-2022-rewire}
Zaiqiao Meng, Fangyu Liu, Ehsan Shareghi, Yixuan Su, Charlotte Collins, and
  Nigel Collier. 2022{\natexlab{b}}.
\newblock \href {https://doi.org/10.18653/v1/2022.acl-long.329}
  {Rewire-then-probe: A contrastive recipe for probing biomedical knowledge of
  pre-trained language models}.
\newblock In \emph{Proceedings of the 60th Annual Meeting of the Association
  for Computational Linguistics (Volume 1: Long Papers)}, pages 4798--4810,
  Dublin, Ireland. Association for Computational Linguistics.

\bibitem[{Mikolov et~al.(2013)Mikolov, Chen, Corrado, and
  Dean}]{mikolov-etal-2013-efficient}
Tomas Mikolov, Kai Chen, Greg Corrado, and Jeffrey Dean. 2013.
\newblock Efficient estimation of word representations in vector space.
\newblock \emph{arXiv preprint arXiv:1301.3781}.

\bibitem[{Min et~al.(2023)Min, Shi, Lewis, Chen, Yih, Hajishirzi, and
  Zettlemoyer}]{min-etal-2023-nonparametric}
Sewon Min, Weijia Shi, Mike Lewis, Xilun Chen, Wen-tau Yih, Hannaneh
  Hajishirzi, and Luke Zettlemoyer. 2023.
\newblock \href {https://doi.org/10.18653/v1/2023.findings-acl.132}
  {Nonparametric masked language modeling}.
\newblock In \emph{Findings of the Association for Computational Linguistics:
  ACL 2023}, pages 2097--2118, Toronto, Canada. Association for Computational
  Linguistics.

\bibitem[{Misra et~al.(2020)Misra, Ettinger, and
  Rayz}]{misra-etal-2020-exploring}
Kanishka Misra, Allyson Ettinger, and Julia Rayz. 2020.
\newblock \href {https://doi.org/10.18653/v1/2020.findings-emnlp.415}
  {Exploring {BERT}{'}s sensitivity to lexical cues using tests from semantic
  priming}.
\newblock In \emph{Findings of the Association for Computational Linguistics:
  EMNLP 2020}, pages 4625--4635, Online. Association for Computational
  Linguistics.

\bibitem[{Newman et~al.(2022)Newman, Choubey, and
  Rajani}]{newman-etal-2022-padapters}
Benjamin Newman, Prafulla~Kumar Choubey, and Nazneen Rajani. 2022.
\newblock \href {https://openreview.net/forum?id=DhzIU48OcZh} {P-{A}dapters:
  Robustly extracting factual information from language models with diverse
  prompts}.
\newblock In \emph{International Conference on Learning Representations}.

\bibitem[{Nye et~al.(2018)Nye, Li, Patel, Yang, Marshall, Nenkova, and
  Wallace}]{nye-etal-2018-corpus}
Benjamin Nye, Junyi~Jessy Li, Roma Patel, Yinfei Yang, Iain Marshall, Ani
  Nenkova, and Byron Wallace. 2018.
\newblock \href {https://doi.org/10.18653/v1/P18-1019} {A corpus with
  multi-level annotations of patients, interventions and outcomes to support
  language processing for medical literature}.
\newblock In \emph{Proceedings of the 56th Annual Meeting of the Association
  for Computational Linguistics (Volume 1: Long Papers)}, pages 197--207,
  Melbourne, Australia. Association for Computational Linguistics.

\bibitem[{Onoe et~al.(2022)Onoe, Zhang, Choi, and
  Durrett}]{onoe-etal-2022-entity}
Yasumasa Onoe, Michael Zhang, Eunsol Choi, and Greg Durrett. 2022.
\newblock \href {https://doi.org/10.18653/v1/2022.findings-naacl.52} {Entity
  cloze by date: What {LM}s know about unseen entities}.
\newblock In \emph{Findings of the Association for Computational Linguistics:
  NAACL 2022}, pages 693--702, Seattle, United States. Association for
  Computational Linguistics.

\bibitem[{OpenAI(2023)}]{openai-2023-gpt4}
OpenAI. 2023.
\newblock \href {http://arxiv.org/abs/2303.08774} {{GPT}-4 technical report}.

\bibitem[{Pandia and Ettinger(2021)}]{pandia-ettinger-2021-sorting}
Lalchand Pandia and Allyson Ettinger. 2021.
\newblock \href {https://doi.org/10.18653/v1/2021.emnlp-main.119} {Sorting
  through the noise: Testing robustness of information processing in
  pre-trained language models}.
\newblock In \emph{Proceedings of the 2021 Conference on Empirical Methods in
  Natural Language Processing}, pages 1583--1596, Online and Punta Cana,
  Dominican Republic. Association for Computational Linguistics.

\bibitem[{Park et~al.(2023)Park, Georgiev, Ilyas, Leclerc, and
  Madry}]{park-etal-2023-trak}
Sung~Min Park, Kristian Georgiev, Andrew Ilyas, Guillaume Leclerc, and
  Aleksander Madry. 2023.
\newblock Trak: Attributing model behavior at scale.
\newblock In \emph{International Conference on Machine Learning (ICML)}.

\bibitem[{Penedo et~al.(2023)Penedo, Malartic, Hesslow, Cojocaru, Cappelli,
  Alobeidli, Pannier, Almazrouei, and Launay}]{penedo-etal-2023-falcon}
Guilherme Penedo, Quentin Malartic, Daniel Hesslow, Ruxandra Cojocaru,
  Alessandro Cappelli, Hamza Alobeidli, Baptiste Pannier, Ebtesam Almazrouei,
  and Julien Launay. 2023.
\newblock The refinedweb dataset for {F}alcon {LLM}: Outperforming curated
  corpora with web data, and web data only.
\newblock \emph{arXiv preprint arXiv:2306.01116}.

\bibitem[{Petroni et~al.(2019)Petroni, Rockt{\"a}schel, Riedel, Lewis, Bakhtin,
  Wu, and Miller}]{petroni-etal-2019-language}
Fabio Petroni, Tim Rockt{\"a}schel, Sebastian Riedel, Patrick Lewis, Anton
  Bakhtin, Yuxiang Wu, and Alexander Miller. 2019.
\newblock \href {https://doi.org/10.18653/v1/D19-1250} {Language models as
  knowledge bases?}
\newblock In \emph{Proceedings of the 2019 Conference on Empirical Methods in
  Natural Language Processing and the 9th International Joint Conference on
  Natural Language Processing (EMNLP-IJCNLP)}, pages 2463--2473, Hong Kong,
  China. Association for Computational Linguistics.

\bibitem[{Podkorytov et~al.(2021)Podkorytov, Biś, and
  Liu}]{podkorytov-etal-2021-limitations}
Maksim Podkorytov, Daniel Biś, and Xiuwen Liu. 2021.
\newblock \href {https://doi.org/10.1109/IJCNN52387.2021.9534299} {How can the
  {[MASK]} know? the sources and limitations of knowledge in bert}.
\newblock In \emph{2021 International Joint Conference on Neural Networks
  (IJCNN)}, pages 1--8.

\bibitem[{Poerner et~al.(2020)Poerner, Waltinger, and
  Sch{\"u}tze}]{poerner-etal-2020-e}
Nina Poerner, Ulli Waltinger, and Hinrich Sch{\"u}tze. 2020.
\newblock \href {https://doi.org/10.18653/v1/2020.findings-emnlp.71}
  {{E}-{BERT}: Efficient-yet-effective entity embeddings for {BERT}}.
\newblock In \emph{Findings of the Association for Computational Linguistics:
  EMNLP 2020}, pages 803--818, Online. Association for Computational
  Linguistics.

\bibitem[{Qin and Eisner(2021)}]{qin-eisner-2021-learning}
Guanghui Qin and Jason Eisner. 2021.
\newblock \href {https://doi.org/10.18653/v1/2021.naacl-main.410} {Learning how
  to ask: Querying {LM}s with mixtures of soft prompts}.
\newblock In \emph{Proceedings of the 2021 Conference of the North American
  Chapter of the Association for Computational Linguistics: Human Language
  Technologies}, pages 5203--5212, Online. Association for Computational
  Linguistics.

\bibitem[{Raffel et~al.(2020)Raffel, Shazeer, Roberts, Lee, Narang, Matena,
  Zhou, Li, and Liu}]{raffel-etal-2020-exploring}
Colin Raffel, Noam Shazeer, Adam Roberts, Katherine Lee, Sharan Narang, Michael
  Matena, Yanqi Zhou, Wei Li, and Peter~J. Liu. 2020.
\newblock \href {http://jmlr.org/papers/v21/20-074.html} {Exploring the limits
  of transfer learning with a unified text-to-text transformer}.
\newblock \emph{Journal of Machine Learning Research}, 21(140):1--67.

\bibitem[{Rajpurkar et~al.(2018)Rajpurkar, Jia, and
  Liang}]{rajpurkar-etal-2018-know}
Pranav Rajpurkar, Robin Jia, and Percy Liang. 2018.
\newblock \href {https://doi.org/10.18653/v1/P18-2124} {Know what you don{'}t
  know: Unanswerable questions for {SQ}u{AD}}.
\newblock In \emph{Proceedings of the 56th Annual Meeting of the Association
  for Computational Linguistics (Volume 2: Short Papers)}, pages 784--789,
  Melbourne, Australia. Association for Computational Linguistics.

\bibitem[{Ray(2023)}]{ray-2023-chatgpt}
Partha~Pratim Ray. 2023.
\newblock \href {https://doi.org/https://doi.org/10.1016/j.iotcps.2023.04.003}
  {Chat{GPT}: A comprehensive review on background, applications, key
  challenges, bias, ethics, limitations and future scope}.
\newblock \emph{Internet of Things and Cyber-Physical Systems}, 3:121--154.

\bibitem[{Roberts et~al.(2020)Roberts, Raffel, and
  Shazeer}]{roberts-etal-2020-much}
Adam Roberts, Colin Raffel, and Noam Shazeer. 2020.
\newblock \href {https://doi.org/10.18653/v1/2020.emnlp-main.437} {How much
  knowledge can you pack into the parameters of a language model?}
\newblock In \emph{Proceedings of the 2020 Conference on Empirical Methods in
  Natural Language Processing (EMNLP)}, pages 5418--5426, Online. Association
  for Computational Linguistics.

\bibitem[{Rogers et~al.(2020)Rogers, Kovaleva, and
  Rumshisky}]{rogers-etal-2020-primer}
Anna Rogers, Olga Kovaleva, and Anna Rumshisky. 2020.
\newblock \href {https://doi.org/10.1162/tacl_a_00349} {A primer in
  {BERT}ology: What we know about how {BERT} works}.
\newblock \emph{Transactions of the Association for Computational Linguistics},
  8:842--866.

\bibitem[{Sadeq et~al.(2022)Sadeq, Xu, and McAuley}]{sadeq-etal-2022-informask}
Nafis Sadeq, Canwen Xu, and Julian McAuley. 2022.
\newblock \href {https://doi.org/10.18653/v1/2022.emnlp-main.395}
  {{I}nfor{M}ask: Unsupervised informative masking for language model
  pretraining}.
\newblock In \emph{Proceedings of the 2022 Conference on Empirical Methods in
  Natural Language Processing}, pages 5866--5878, Abu Dhabi, United Arab
  Emirates. Association for Computational Linguistics.

\bibitem[{Saeed and Papotti(2022)}]{saeed-papotti-2022-type}
Mohammed Saeed and Paolo Papotti. 2022.
\newblock \href {https://doi.org/10.18653/v1/2022.findings-emnlp.336} {You are
  my type! type embeddings for pre-trained language models}.
\newblock In \emph{Findings of the Association for Computational Linguistics:
  EMNLP 2022}, pages 4583--4598, Abu Dhabi, United Arab Emirates. Association
  for Computational Linguistics.

\bibitem[{Safavi and Koutra(2021)}]{safavi-koutra-2021-relational}
Tara Safavi and Danai Koutra. 2021.
\newblock \href {https://doi.org/10.18653/v1/2021.emnlp-main.81} {{R}elational
  {W}orld {K}nowledge {R}epresentation in {C}ontextual {L}anguage {M}odels: {A}
  {R}eview}.
\newblock In \emph{Proceedings of the 2021 Conference on Empirical Methods in
  Natural Language Processing}, pages 1053--1067, Online and Punta Cana,
  Dominican Republic. Association for Computational Linguistics.

\bibitem[{Sciavolino et~al.(2021)Sciavolino, Zhong, Lee, and
  Chen}]{sciavolino-etal-2021-simple}
Christopher Sciavolino, Zexuan Zhong, Jinhyuk Lee, and Danqi Chen. 2021.
\newblock \href {https://doi.org/10.18653/v1/2021.emnlp-main.496} {Simple
  entity-centric questions challenge dense retrievers}.
\newblock In \emph{Proceedings of the 2021 Conference on Empirical Methods in
  Natural Language Processing}, pages 6138--6148, Online and Punta Cana,
  Dominican Republic. Association for Computational Linguistics.

\bibitem[{Shi et~al.(2021)Shi, Joshi, and Zettlemoyer}]{shi-etal-2021-descgen}
Weijia Shi, Mandar Joshi, and Luke Zettlemoyer. 2021.
\newblock \href {https://doi.org/10.18653/v1/2021.acl-long.35} {{DESCGEN}: A
  distantly supervised datasetfor generating entity descriptions}.
\newblock In \emph{Proceedings of the 59th Annual Meeting of the Association
  for Computational Linguistics and the 11th International Joint Conference on
  Natural Language Processing (Volume 1: Long Papers)}, pages 415--427, Online.
  Association for Computational Linguistics.

\bibitem[{Shin et~al.(2020)Shin, Razeghi, Logan~IV, Wallace, and
  Singh}]{shin-etal-2020-autoprompt}
Taylor Shin, Yasaman Razeghi, Robert~L. Logan~IV, Eric Wallace, and Sameer
  Singh. 2020.
\newblock \href {https://doi.org/10.18653/v1/2020.emnlp-main.346}
  {{A}uto{P}rompt: {E}liciting {K}nowledge from {L}anguage {M}odels with
  {A}utomatically {G}enerated {P}rompts}.
\newblock In \emph{Proceedings of the 2020 Conference on Empirical Methods in
  Natural Language Processing (EMNLP)}, pages 4222--4235, Online. Association
  for Computational Linguistics.

\bibitem[{Singhania et~al.(2022)Singhania, Nguyen, and
  Razniewski}]{singhania-etal-2022-knowledge}
Sneha Singhania, Tuan-Phong Nguyen, and Simon Razniewski. 2022.
\newblock Knowledge base construction from pre-trained language models 2022.
\newblock In \emph{Semantic Web Challenge on Knowledge Base Construction from
  Pre-trained Language Models}. CEUR-WS.

\bibitem[{Sung et~al.(2021)Sung, Lee, Yi, Jeon, Kim, and
  Kang}]{sung-etal-2021-language}
Mujeen Sung, Jinhyuk Lee, Sean Yi, Minji Jeon, Sungdong Kim, and Jaewoo Kang.
  2021.
\newblock \href {https://doi.org/10.18653/v1/2021.emnlp-main.388} {Can language
  models be biomedical knowledge bases?}
\newblock In \emph{Proceedings of the 2021 Conference on Empirical Methods in
  Natural Language Processing}, pages 4723--4734, Online and Punta Cana,
  Dominican Republic. Association for Computational Linguistics.

\bibitem[{Taylor(1953)}]{taylor-1953-cloze}
Wilson~L. Taylor. 1953.
\newblock \href {https://doi.org/10.1177/107769905303000401} {“{C}loze
  {P}rocedure”: A new tool for measuring readability}.
\newblock \emph{Journalism Quarterly}, 30(4):415--433.

\bibitem[{Tenney et~al.(2019)Tenney, Xia, Chen, Wang, Poliak, McCoy, Kim,
  Durme, Bowman, Das, and Pavlick}]{tenney-etal-2018-context}
Ian Tenney, Patrick Xia, Berlin Chen, Alex Wang, Adam Poliak, R~Thomas McCoy,
  Najoung Kim, Benjamin~Van Durme, Sam Bowman, Dipanjan Das, and Ellie Pavlick.
  2019.
\newblock \href {https://openreview.net/forum?id=SJzSgnRcKX} {What do you learn
  from context? probing for sentence structure in contextualized word
  representations}.
\newblock In \emph{International Conference on Learning Representations}.

\bibitem[{Touvron et~al.(2023)Touvron, Lavril, Izacard, Martinet, Lachaux,
  Lacroix, Rozi{\`e}re, Goyal, Hambro, Azhar, Rodriguez, Joulin, Grave, and
  Lample}]{touvron-etal-2023-llama}
Hugo Touvron, Thibaut Lavril, Gautier Izacard, Xavier Martinet, Marie-Anne
  Lachaux, Timoth{\'e}e Lacroix, Baptiste Rozi{\`e}re, Naman Goyal, Eric
  Hambro, Faisal Azhar, Aurelien Rodriguez, Armand Joulin, Edouard Grave, and
  Guillaume Lample. 2023.
\newblock {LL}a{MA}: Open and efficient foundation language models.
\newblock \emph{arXiv preprint arXiv:2302.13971}.

\bibitem[{Wallat et~al.(2020)Wallat, Singh, and
  Anand}]{singh-etal-2020-bertnesia}
Jonas Wallat, Jaspreet Singh, and Avishek Anand. 2020.
\newblock \href {https://doi.org/10.18653/v1/2020.blackboxnlp-1.17}
  {{BERT}nesia: Investigating the capture and forgetting of knowledge in
  {BERT}}.
\newblock In \emph{Proceedings of the Third BlackboxNLP Workshop on Analyzing
  and Interpreting Neural Networks for NLP}, pages 174--183, Online.
  Association for Computational Linguistics.

\bibitem[{Wang et~al.(2021{\natexlab{a}})Wang, Liu, and
  Zhang}]{wang-etal-2021-generative}
Cunxiang Wang, Pai Liu, and Yue Zhang. 2021{\natexlab{a}}.
\newblock \href {https://doi.org/10.18653/v1/2021.acl-long.251} {Can generative
  pre-trained language models serve as knowledge bases for closed-book {QA}?}
\newblock In \emph{Proceedings of the 59th Annual Meeting of the Association
  for Computational Linguistics and the 11th International Joint Conference on
  Natural Language Processing (Volume 1: Long Papers)}, pages 3241--3251,
  Online. Association for Computational Linguistics.

\bibitem[{Wang et~al.(2021{\natexlab{b}})Wang, Tang, Duan, Wei, Huang, Ji, Cao,
  Jiang, and Zhou}]{wang-etal-2021-k}
Ruize Wang, Duyu Tang, Nan Duan, Zhongyu Wei, Xuanjing Huang, Jianshu Ji,
  Guihong Cao, Daxin Jiang, and Ming Zhou. 2021{\natexlab{b}}.
\newblock \href {https://doi.org/10.18653/v1/2021.findings-acl.121}
  {{K-Adapter}: {I}nfusing {K}nowledge into {P}re-{T}rained {M}odels with
  {A}dapters}.
\newblock In \emph{Findings of the Association for Computational Linguistics:
  ACL-IJCNLP 2021}, pages 1405--1418, Online. Association for Computational
  Linguistics.

\bibitem[{Wang et~al.(2023)Wang, Lu, Kong, and
  Sang}]{wang-etal-2023-towards-alleviating}
Yuhang Wang, Dongyuan Lu, Chao Kong, and Jitao Sang. 2023.
\newblock \href {https://doi.org/10.18653/v1/2023.findings-acl.270} {Towards
  alleviating the object bias in prompt tuning-based factual knowledge
  extraction}.
\newblock In \emph{Findings of the Association for Computational Linguistics:
  ACL 2023}, pages 4420--4432, Toronto, Canada. Association for Computational
  Linguistics.

\bibitem[{Whitehouse et~al.(2022)Whitehouse, Christopoulou, and
  Iacobacci}]{whitehouse-etal-2022-entitycs}
Chenxi Whitehouse, Fenia Christopoulou, and Ignacio Iacobacci. 2022.
\newblock \href {https://doi.org/10.18653/v1/2022.findings-emnlp.499}
  {{E}ntity{CS}: Improving zero-shot cross-lingual transfer with entity-centric
  code switching}.
\newblock In \emph{Findings of the Association for Computational Linguistics:
  EMNLP 2022}, pages 6698--6714, Abu Dhabi, United Arab Emirates. Association
  for Computational Linguistics.

\bibitem[{Xiong et~al.(2020)Xiong, Du, Wang, and
  Stoyanov}]{xiong-etal-2020-pretrained}
Wenhan Xiong, Jingfei Du, William~Yang Wang, and Veselin Stoyanov. 2020.
\newblock \href {https://openreview.net/forum?id=BJlzm64tDH} {Pretrained
  encyclopedia: Weakly supervised knowledge-pretrained language model}.
\newblock In \emph{International Conference on Learning Representations}.

\bibitem[{Ye et~al.(2022)Ye, Gao, Li, Xu, Feng, Wu, Yu, and
  Kong}]{ye-etal-2022-zerogen}
Jiacheng Ye, Jiahui Gao, Qintong Li, Hang Xu, Jiangtao Feng, Zhiyong Wu, Tao
  Yu, and Lingpeng Kong. 2022.
\newblock \href {https://doi.org/10.18653/v1/2022.emnlp-main.801} {{Z}ero{G}en:
  Efficient zero-shot learning via dataset generation}.
\newblock In \emph{Proceedings of the 2022 Conference on Empirical Methods in
  Natural Language Processing}, pages 11653--11669, Abu Dhabi, United Arab
  Emirates. Association for Computational Linguistics.

\bibitem[{Ye et~al.(2021)Ye, Li, Wang, Bolte, Ma, Yih, Ren, and
  Khabsa}]{ye-etal-2021-influence}
Qinyuan Ye, Belinda~Z. Li, Sinong Wang, Benjamin Bolte, Hao Ma, Wen-tau Yih,
  Xiang Ren, and Madian Khabsa. 2021.
\newblock \href {https://doi.org/10.18653/v1/2021.emnlp-main.573} {On the
  influence of masking policies in intermediate pre-training}.
\newblock In \emph{Proceedings of the 2021 Conference on Empirical Methods in
  Natural Language Processing}, pages 7190--7202, Online and Punta Cana,
  Dominican Republic. Association for Computational Linguistics.

\bibitem[{Yoshikawa and Okazaki(2023)}]{yoshikawa-okazaki-2023-selective}
Hiyori Yoshikawa and Naoaki Okazaki. 2023.
\newblock \href {https://doi.org/10.18653/v1/2023.findings-eacl.150}
  {Selective-{LAMA}: Selective prediction for confidence-aware evaluation of
  language models}.
\newblock In \emph{Findings of the Association for Computational Linguistics:
  EACL 2023}, pages 2017--2028, Dubrovnik, Croatia. Association for
  Computational Linguistics.

\bibitem[{Yuan et~al.(2021)Yuan, Liu, Tan, Huang, and
  Huang}]{yuan-etal-2021-improving}
Zheng Yuan, Yijia Liu, Chuanqi Tan, Songfang Huang, and Fei Huang. 2021.
\newblock \href {https://doi.org/10.18653/v1/2021.bionlp-1.20} {Improving
  biomedical pretrained language models with knowledge}.
\newblock In \emph{Proceedings of the 20th Workshop on Biomedical Language
  Processing}, pages 180--190, Online. Association for Computational
  Linguistics.

\bibitem[{Zhang and Choi(2021)}]{zhang-choi-2021-situatedqa}
Michael Zhang and Eunsol Choi. 2021.
\newblock \href {https://doi.org/10.18653/v1/2021.emnlp-main.586}
  {{S}ituated{QA}: Incorporating extra-linguistic contexts into {QA}}.
\newblock In \emph{Proceedings of the 2021 Conference on Empirical Methods in
  Natural Language Processing}, pages 7371--7387, Online and Punta Cana,
  Dominican Republic. Association for Computational Linguistics.

\bibitem[{Zhang et~al.(2021)Zhang, Warstadt, Li, and
  Bowman}]{zhang-etal-2021-need}
Yian Zhang, Alex Warstadt, Xiaocheng Li, and Samuel~R. Bowman. 2021.
\newblock \href {https://doi.org/10.18653/v1/2021.acl-long.90} {When do you
  need billions of words of pretraining data?}
\newblock In \emph{Proceedings of the 59th Annual Meeting of the Association
  for Computational Linguistics and the 11th International Joint Conference on
  Natural Language Processing (Volume 1: Long Papers)}, pages 1112--1125,
  Online. Association for Computational Linguistics.

\bibitem[{Zhang et~al.(2022)Zhang, Fei, Li, and Li}]{zhang-etal-2022-promptgen}
Yue Zhang, Hongliang Fei, Dingcheng Li, and Ping Li. 2022.
\newblock \href {https://doi.org/10.18653/v1/2022.findings-naacl.3}
  {{P}rompt{G}en: Automatically generate prompts using generative models}.
\newblock In \emph{Findings of the Association for Computational Linguistics:
  NAACL 2022}, pages 30--37, Seattle, United States. Association for
  Computational Linguistics.

\bibitem[{Zhao et~al.(2021)Zhao, Wallace, Feng, Klein, and
  Singh}]{zhao-etal-2021-calibrate}
Zihao Zhao, Eric Wallace, Shi Feng, Dan Klein, and Sameer Singh. 2021.
\newblock \href {https://proceedings.mlr.press/v139/zhao21c.html} {Calibrate
  before use: Improving few-shot performance of language models}.
\newblock In \emph{Proceedings of the 38th International Conference on Machine
  Learning}, volume 139 of \emph{Proceedings of Machine Learning Research},
  pages 12697--12706. PMLR.

\bibitem[{Zhong et~al.(2023)Zhong, Ding, Liu, Du, and
  Tao}]{zhong-etal-2023-self}
Qihuang Zhong, Liang Ding, Juhua Liu, Bo~Du, and Dacheng Tao. 2023.
\newblock \href {https://doi.org/10.18653/v1/2023.findings-acl.254}
  {Self-evolution learning for discriminative language model pretraining}.
\newblock In \emph{Findings of the Association for Computational Linguistics:
  ACL 2023}, pages 4130--4145, Toronto, Canada. Association for Computational
  Linguistics.

\bibitem[{Zhong et~al.(2021)Zhong, Friedman, and
  Chen}]{zhong-etal-2021-factual}
Zexuan Zhong, Dan Friedman, and Danqi Chen. 2021.
\newblock \href {https://doi.org/10.18653/v1/2021.naacl-main.398} {Factual
  probing is [{MASK}]: Learning vs. learning to recall}.
\newblock In \emph{Proceedings of the 2021 Conference of the North American
  Chapter of the Association for Computational Linguistics: Human Language
  Technologies}, pages 5017--5033, Online. Association for Computational
  Linguistics.

\bibitem[{Zini and Awad(2022)}]{zini-etal-2022-explainability}
Julia~El Zini and Mariette Awad. 2022.
\newblock \href {https://doi.org/10.1145/3529755} {On the explainability of
  natural language processing deep models}.
\newblock \emph{ACM Comput. Surv.}, 55(5).

\end{thebibliography}
\bibliographystyle{acl_natbib}


\appendix
\section{Corpus Creation \& Annotation Methodology}
\subsection{Paper Selection}
\label{app:annotationsummary}
We created our corpus of papers about factual knowledge probing by querying the Semantic Scholar API\footnote{\url{https://api.semanticscholar.org/graph/v1/paper/d0086b86103a620a86bc918746df0aa642e2a8a3/citations?fields=intents,url,title,abstract,venue,year,referenceCount,citationCount,influentialCitationCount,fieldsOfStudy,publicationDate&limit=1000}} on 31. August, 2023 at 2:51 pm for all works citing~\cite{petroni-etal-2019-language}, resulting in 1,416 papers. According to our knowledge~\citet{petroni-etal-2019-language}'s work was the first to quantify factual knowledge in PLMs, and envisioned using PLMs as KBs. We then separated the papers based on their venue information: the venue papers (1,006 instances), the arXiv papers (375 instances) and the no-venue papers (35 instances) with missing venue information returned by the Semantic Scholar API. We did a web search for the no-venue papers and manually assigned them to their respective group, yielding a total of 1,008 venue papers. We then proceeded to annotate in two steps, after which our final corpus of papers were reduced to 94 highly relevant papers.
\subsection{First Annotation Step}
\label{app:annotation}
Three authors single annotated the peer-reviewed papers as either relevant or not-relevant based on our inclusion and exclusion criteria, discussing uncertainties with all annotators. When in doubt, the paper was marked as relevant for a high recall. 

A paper was marked as relevant based on its title, abstract or body, if it contained one of the following:
\begin{itemize}[nosep, leftmargin=*]
    \item Knowledge probing methods, metrics or datasets for quantifying the relational or factual knowledge stored in a PLM.
    \item Methods shown to increase the amount of relational or factual knowledge stored in a PLM. 
    \item Methods to update, localize or increase the consistency of factual knowledge in PLMs.
    
\end{itemize}

 We further explicitly included short and long papers, workshop papers, posters, shared tasks, competitions and challenge papers with at least 4 pages and explicitly excluded master theses, PhD dissertations, workshop proposals, workshop reports and other non-peer-reviewed publications. After deduplication, the first annotation step yielded a total of 173 relevant papers. 
\subsection{Second Annotation Step}
\label{app:second_annotation_step}
\begin{table}
    \centering
    \begin{tabular}{+c^p{6.75cm}}
    \toprule
    1 & Which probing methods are used? \\
    2 & Which probing methods are proposed? \\
    3 & Which PLMs are probed? \\
    4 & Are the probed PLMs fine-tuned? \\
    5 & Which knowledge probing datasets are used? \\
    6 & What are the sources for the datasets? \\
    7 & Does the paper present any methods or analysis with respect to factual knowledge probing (e.g., updability, interpretability, consistency)? \\
    8 & Does the paper propose a method to inject knowledge into a PLM? \\
    9 & Is the paper a survey? \\\bottomrule
    \end{tabular}
    \caption{The information extracted for each paper.}
    \label{tab:paper-annotation-questions}
\end{table}


We revisited all relevant papers and annotated them based on the questions listed in table~\ref{tab:paper-annotation-questions}. We only included publications that perform intrinsic evaluations~\cite{kalyan-etal-2021-ammus} that directly target factual knowledge, and we exclude extrinsic evaluations on knowledge-intensive tasks. Additionally, we excluded papers that did not train on free-text corpora. For each probing method, we store its name, description and an example. For datasets, we store name, domain (of knowledge), original source (from which the dataset was compiled), number of instances and language of the dataset. We excluded 88 of 173 papers for being irrelevant upon second inspection and added 9 papers, which were not in our initial corpus of research papers. These papers introduced relevant datasets. Our final corpus of surveyed papers thus counts 94 highly relevant papers.

\begin{table*}
\begin{minipage}{\linewidth}
\tiny
\centering
\resizebox{\textwidth}{!}{
\begin{tabular}{+l^>{\raggedright\hangindent=0.5em}p{4.4cm}^l^p{0.7cm}^p{5.2cm}^c^c}

\toprule \tabhead
& Dataset &  Cat. &   Lang. &  Example & \#Inst.  & Access \\
\otoprule
\multirow{20}{*}{\rotatebox[origin=c]{90}{\sc General knowledge}}
& {\cmss LAMA}~\cite{petroni-etal-2019-language} 
    & \gk \cloze
    & en  
    & \prex{Dante was born in [MASK]} 
    & 40k  
    & + \\ 
& \cite{bouraoui-etal-2020-inducing} {\cmss Google Analogy (semantic)}~\cite{mikolov-etal-2013-efficient} 
    & \gk \cls 
    & en 
    &  \prex{It is located in [X], the capital of [Y]}  
    & 9k 
    & + \\                       
& \cite{roberts-etal-2020-much} {\cmss WebQuestions}~\cite{berant-etal-2013-semantic} 
    & \gk \questions 
    & en  
    & \prex{What degrees did Obama get?} 
    & 6k 
    & + \\         
& \cite{bouraoui-etal-2020-inducing} {\cmss BATS (encyclopedic)} ~\cite{gladkova-etal-2016-analogy} 
    & \gk \cls 
    & en 
    & \prex{[X] is the capital of [Y]} 
    & 0.5k 
    & + \\                            
& \cite{roberts-etal-2020-much} {\cmss TriviaQA}~\cite{joshi-etal-2017-triviaqa}
    & \gk \questions 
    & en  
    & \prex{Who won the Nobel Peace Prize in 2009?} 
    & 96k 
    & + \\              
& \cite{roberts-etal-2020-much} {\cmss NQ}~\cite{kwiatkowski-etal-2019-natural} 
    & \gk \questions
    & en  
    & \prex{Who lives in the imperial palace in Tokyo?} 
    & 322k 
    & + \\
& {\cmss IndicGLUE}~\cite{kakwani-etal-2020-indicnlpsuite} 
    & \gk \cloze 
    & indic\footnote{11 different languages}
    & \prex{Shambhupara <MASK> is an important village in Amreli Tehsil, Gujarat State.}
    & 239k 
    & + \\ 
& {\cmss X-FACTR}~\cite{jiang-etal-2020-x} 
    & \gk \cloze
    & multi 
    & \prex{The mother tongue of Obama is [MASK]} 
    & 398k 
    & + \\
& {\cmss LAMA-UHN} ~\cite{poerner-etal-2020-e}    
    & \gk \cloze 
    & en  
    & \prex{USA maintains diplomatic relations with [MASK]} 
    & 32k  
    & o \\ 
& {\cmss LPAQA}~\cite{jiang-etal-2020-know} 
    & \gk \cloze
    & en 
    & \prex{DirectX is developed/created by [MASK]} 
    & 3k 
    & + \\ 
& {\cmss mLAMA}~\cite{kassner-etal-2021-multilingual} 
    & \gk \cloze 
    & multi 
    & \prex{Paris is the capital of [MASK]} 
    & 855k 
    & + \\
& {\cmss DESCGEN}~\cite{shi-etal-2021-descgen} 
    & \gk \nlg 
    & en 
    & \prex{[Carl Menger] was an Austrian economist...} 
    & 37k 
    & + \\
& {\cmss WIKI-UNI}~\cite{cao-etal-2021-knowledgeable} 
    & \gk \cloze 
    & en 
    & \prex{Turing was born in [MASK].} 
    & 70k 
    & + \\
& \cite{wang-etal-2021-generative} \cite{rajpurkar-etal-2018-know}
   & \gk \questions
   & en
   & \prex{<Q> $\rightarrow$ <answer related passage> <A>}
   & 92k
   & + \\

& {\cmss KAMEL}~\cite{kalo-fichtel-2022-kamel} 
    & \gk \questions 
    & en 
    & \prex{<Q\&A>*, What languages does Confuzius speak?} 
    & 47k 
    & + \\
& \CR{{\cmss DLAMA}~\cite{keleg-magdy-2023-dlama}}
    & \gk \cloze
    & multi
    & \prex{Egypt is located in [MASK]} 
    & 78k
    & + \\
& \CR{{\cmss PopQA}~\cite{mallen-etal-2023-trust}}
    & \gk \questions 
    & en 
    & \prex{What is the capital of Louisiana?} 
    & 14K 
    & + \\

&  \CR{\cite{mallen-etal-2023-trust}{\cmss EntityQuestions}~\cite{sciavolino-etal-2021-simple}} 
    & \gk \questions 
    & en 
    & \prex{Who is the author of The Target?} 
    & 177k
    & + \\
    \midrule
\multirow{9}{*}{\rotatebox[origin=c]{90}{\sc Domain-specific}}
& {\cmss EXAMS}~\cite{hardalov-etal-2020-exams} 
    & \dk \questions
    & multi 
    & \prex{<Q> <A1,A2,A3,A4> $\rightarrow$ <A\textsubscript{i}>} 
    & 24k 
    & + \\
& {\cmss MedQA}~\cite{jin-etal-2021-disease} 
    & \dk \questions
    & en,zh\footnote{including zh-simplified}
    &  \prex{<Case> <Q> <A1,A2,A3,A4> $\rightarrow$ <A\textsubscript{i}>} 
    & 61k 
    & + \\             
& {\cmss DisKnE}~\cite{alghanmi-etal-2021-probing} 
    & \dk \cls
    & en 
    & \prex{The patient has high BP <SEP> Hypertension} 
    & 7k 
    & o \\  
& \cite{yuan-etal-2021-improving} 
    & \dk \cloze 
    & en 
    & \prex{apraclonidine may prevent [MASK]} 
    & 144k 
    & o \\

& {\cmss LEFT}~\cite{ciosici-etal-2021-perhaps} 
    & \dk \cls 
    & en 
    &  \prex{<statement> $\rightarrow$ <True/False>}  
    & 1k 
    & o \\

& {\cmss BioLAMA}~\cite{sung-etal-2021-language} 
    & \dk \cloze
    & en  
    & \prex{Hepatitis has symptoms such as [MASK]} 
    & 49k 
    & + \\ 
& \cite{abaho-etal-2022-position} {\cmss EBM-NLP}~\cite{nye-etal-2018-corpus} 
    & \dk  \cloze
    & en  
    & \prex{...patient showed complete [MASK]} 
    & 3k 
    & - \\ 
& {\cmss MedLAMA}~\cite{meng-etal-2022-rewire} 
    & \dk \cloze  
    & en  
    & \prex{Elvitegravir may prevent [MASK]} 
    & 19k 
    & + \\  \midrule
\multirow{18}{*}{\rotatebox[origin=c]{90}{\sc Other}}
& {\cmss Negated LAMA}~\cite{kassner-schutze-2020-negated} 
    & \consistency \cloze
    & en 
    & \prex{The capital of Italy is not [MASK]} 
    & 10k 
    & + \\
& {\cmss Misprimed LAMA}~\cite{kassner-schutze-2020-negated} 
    & \consistency \cloze
    & en 
    & \prex{Dinosaurs? Munich is located [MASK]} 
    & 11k 
    & + \\
& {\cmss ParaRel}~\cite{elazar-etal-2021-measuring} 
    & \consistency \cloze
    & en 
    & \prex{Turing is from/was born in [MASK]} 
    & n.a.\footnote{number of relations 328, prompts per relation 38}  
    & + \\                              
& {\cmss \cite{pandia-ettinger-2021-sorting}}
    & \consistency \cloze 
    & en
    & \prex{Sebastian lives in France. The capital of Sebastian’s country is [MASK].} 
    & 40k
    & + \\                  
& {\cmss SituatedQA (context/answer)} \cite{zhang-choi-2021-situatedqa} 
    & \contextk \questions  
    & en  
    & \prex{Who made the most 3 point shots in the NBA?} 
    & 18k 
    & + \\ 
& \cite{heinzerling-inui-2021-language} 
    & \kb \cloze
    & en 
    & \prex{Turing was born in [MASK]} 
    & 15M 
    & + \\                               

& \cite{podkorytov-etal-2021-limitations} 
    & \mc \cloze 
    & en 
    & \prex{Tomatoes are a [MASK].} 
    & 0.1k   
    & - \\
& {\cmss mParaRel}~\cite{fierro-sogaard-2022-factual} 
    & \consistency \cloze 
    & multi 
    & \prex{Turing is from/was born in [MASK]}  
    & n.a.\footnote{number of relations 343, prompts per relation 37.13 (avg. over languages)}   
    & + \\
& {\cmss TEMPLAMA}~\cite{dhingra-etal-2022-time} 
    & \contextk  \cloze
    & en  
    & \prex{[2012] Cristiano Ronaldo plays for [MASK]} 
    & 50k 
    & + \\                                                                                       
& \cite{singhania-etal-2022-knowledge} 
    & \kb \cloze 
    & en 
    & \prex{France shares a land border with [MASK]} 
    & 2k 
    & + \\
& \cite{jang-etal-2022-towards} 
    & \ku \cloze
    & en 
    & \prex{[MASK] is the prime minister of England} 
    & 30k 
    & + \\
& \CR{ \cmss{TemporalWiki}~\cite{jang-etal-2022-temporalwiki}}	
    & \ku \ppl
    & en
    & \prex{On 1 December, the Omicron variant...}
    & \textcircled{d} 
    & o \\
& \cite{lee-etal-2022-plug} {\cmss zsRE}~\cite{levy-etal-2017-zero} 
    & \ku \questions 
    & en  
    & \prex{Who is the most paid player in EPL?} 
    & 168k 
    & + \\
& CounterFact~\cite{meng-etal-2022-locating}
   & \ku \cloze
   & en
   & \prex{Turing's mother tongue is <old,new>} 
   & 22k
   & + \\
& {\cmss ECBD}~\cite{onoe-etal-2022-entity} 
    & \ue \cloze 
    & en  
    & \prex{[mRNA vaccines] do not affect [MASK].} 
    & 35k 
    & + \\
& \CR{\cite{hase-etal-2023-methods}}
    & \ku \cloze
    & en 
    & \prex{Mary Lowe Good has relation ‘winner of’ to [MASK]} 
    & 170k 
    & + \\

& \CR{ \cmss{CounterFact+}~\cite{hoelscher-obermaier-etal-2023-detecting} }
    & \ku \cloze 
    & en 
    & \prex{The mother tongue of Danielle Darrieux is English. The native language of Montesquieu is [MASK]} &  \textcircled{d} 
    & o \\
& \CR{\cmss{DynamicTempLAMA}~\cite{margatina-etal-2023-dynamic} }
    & \ku \cloze 
    & en 
    & \prex{The surname of the Prime Minister of the UK is [MASK]}
    & \textcircled{d}
    & + \\ 		
\bottomrule
\end{tabular}
}
\caption{Datasets for factual knowledge probing. 
Probed knowledge: 
 general knowledge \gk, domain-specific knowledge \dk, context-dependent knowledge \contextk, PLMs sensitivity to paraphrases, negation or mispriming \consistency, related to PLMs as KBs \kb{} , knowledge updating \ku ,  misconceptions \mc{} and  unseen entities \ue. NLP task: cloze prompts \cloze, question answering \questions, classification \cls, natural language generation \nlg, \CR{and perplexity} \ppl. \CR{\textcircled{d} refers to dynamic datasets, whose number of instances is changeable over time.}} 
 \label{app:datasets}
\end{minipage}
\end{table*}

\end{document}